\documentclass[11pt]{article}

\newif\ifarxivversion

\arxivversiontrue

\ifarxivversion
    \usepackage[preprint]{acl}
\else
    \usepackage[review]{acl}
\fi

\usepackage{times}
\usepackage{latexsym}
\usepackage{amsmath,amssymb}

\usepackage[T1]{fontenc}

\usepackage[utf8]{inputenc}

\usepackage{microtype}

\usepackage{inconsolata}

\usepackage{graphicx}

\usepackage{booktabs}
\usepackage{graphicx}
\usepackage{multirow}

\title{Support or Magnitude? Characterizing Neuron Relevance for Instruction Data Selection}

\ifarxivversion

\author{
  Yifan Le \\
  Independent Researcher \\
  \texttt{leyifan@zju.edu.cn}
}

\else

\author{}

\fi

\begin{document}
\maketitle
\begin{abstract}
Model-internal signals, such as neuron activations
and gradients, have been used for instruction data
selection, yet how neuron relevance should be
characterized remains underexplored.
In particular, should selection prioritize the
magnitude of relevance or how broadly it is
distributed across neurons?
To investigate this question, we construct
sample-level relevance profiles using AttnLRP-guided
attribution and introduce two complementary criteria:
Effective Relevance Support (ERS), which measures
the effective support of positive relevance, and
Positive Relevance Magnitude (PRM), which captures
its average strength.
Our analyses reveal that these characterizations
produce distinct sample rankings and selection
preferences despite sharing the same underlying
relevance profiles.
On Qwen2 models, PRM exhibits stronger associations
with activation-based selection, whereas ERS
produces substantially different rankings.
These differences extend to task composition,
within-task sample preferences, and semantic
selection patterns, with varying degrees of
agreement across backbones.
Experiments across three language models and five
downstream benchmarks demonstrate the practical
value of relevance-guided selection, with ERS
achieving the best overall average performance
rank among the evaluated methods.
Our findings highlight that how neuron relevance
is distributed and characterized, rather than
its magnitude alone, matters for instruction
data selection.
\end{abstract}


\begin{figure*}[t]
\centering
\includegraphics[
    width=0.98\textwidth
]{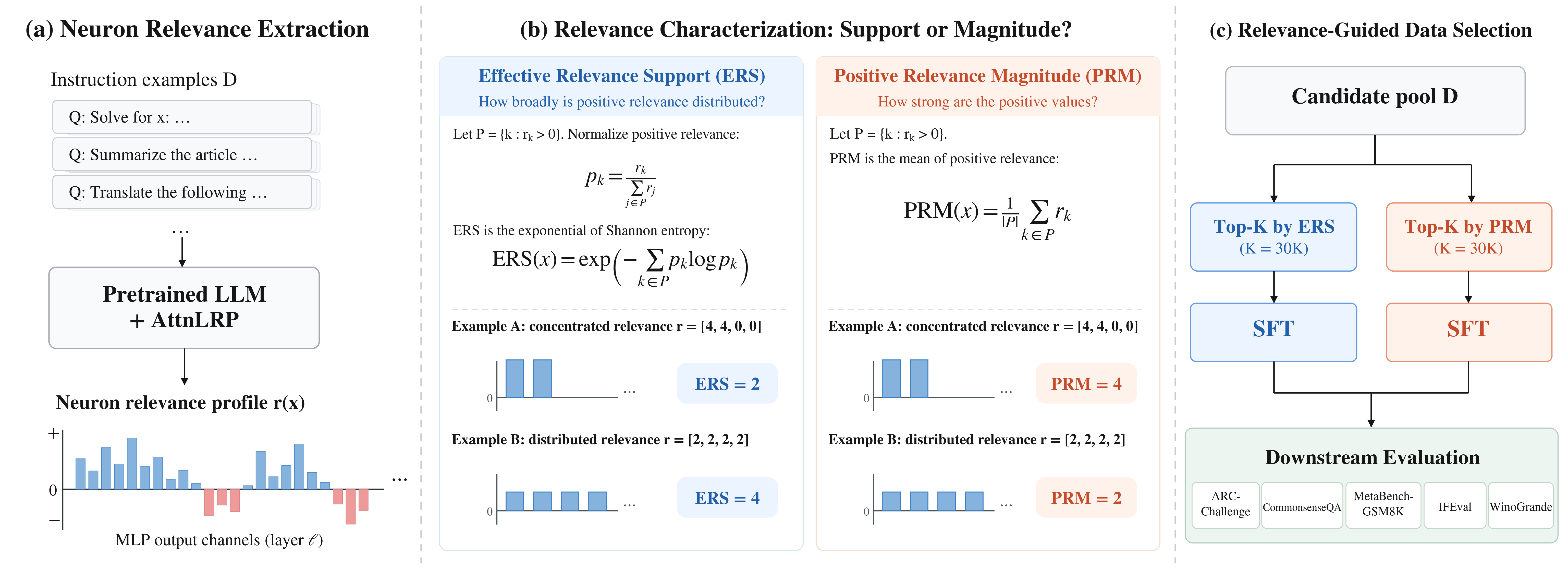}

\caption{
Overview of the proposed neuron relevance-based
instruction data selection framework.
We first compute neuron-level relevance profiles
using AttnLRP for candidate instruction examples.
Two complementary selection criteria are introduced:
Effective Relevance Support (ERS), which captures
the breadth of positive relevance, and
Positive Relevance Magnitude (PRM), which measures
its magnitude.
Examples are ranked and selected for supervised
fine-tuning, followed by evaluation on five
downstream benchmarks.
}
\label{fig:framework_overview}
\end{figure*}

\section{Introduction}
\label{sec:introduction}

Supervised fine-tuning (SFT) is essential for
equipping large language models (LLMs) with
instruction-following capabilities.
Prior studies have shown that carefully selected
instruction subsets can achieve competitive
performance with substantially fewer examples
\citep{zhou2023lima,chen2024alpagasus,liu2024deita}.
Identifying valuable training examples is thus
central to data-efficient instruction tuning.

Existing selection methods rely on data quality,
complexity, diversity, and increasingly,
model-internal signals such as gradients,
neuron activations, and learned features
\citep{lu2024instag,li-etal-2024-quantity,
cao2024instruction,xia2024less,
dai-etal-2025-improving,chen2026neuron,
ma2025monosemantic}.
However, how neuron relevance should be
characterized for data selection remains
largely unexplored.
In particular, \emph{should selection prioritize
relevance magnitude or its effective support
across neurons?}

Unlike activation, which reflects neuronal
responses, relevance provides an attribution-based
view of internal model signals.
Layer-wise Relevance Propagation (LRP)
\citep{bach2015lrp} and AttnLRP
\citep{achtibat2024attnlrp} enable us to
investigate such relevance distributions.
Importantly, different characterizations of
the same relevance profile may lead to
different assessments of training examples.

To investigate this question, we construct
sample-level \emph{Neuron Relevance Profiles}
using AttnLRP-guided attribution and introduce
two complementary selection criteria:
\textbf{Effective Relevance Support (ERS)},
which measures the effective support of positive
relevance, and
\textbf{Positive Relevance Magnitude (PRM)},
which captures its average strength.
Both operate on the same relevance profiles,
allowing us to examine how different
characterizations affect data selection.

Experiments across three language models and
five downstream benchmarks reveal that ERS
and PRM induce distinct rankings and selection
preferences.
On Qwen2 models, PRM is more strongly associated
with activation-based selection, whereas ERS
produces substantially different rankings.
These differences extend to within-task selection
and vary across model backbones.
Furthermore, \textbf{ERS achieves the best overall
average performance rank} among the evaluated
methods.
Figure~\ref{fig:framework_overview}
illustrates our framework.

Our contributions are threefold:

\textbf{Relevance characterization.}
We introduce ERS and PRM to characterize
the effective support and average magnitude
of neuron relevance for instruction
data selection.

\textbf{Empirical effectiveness.}
Across three models and five benchmarks,
ERS achieves the best overall average
performance rank among evaluated methods.

\textbf{Selection behavior analysis.}
We demonstrate that different relevance
characterizations produce distinct
rankings, task-conditioned preferences,
and backbone-dependent selection patterns.

\section{Related Work}
\label{sec:related-work}

\subsection{Sample-Centric Data Selection}
\label{sec:rw-sample}

Instruction tuning depends critically on data quality,
composition, and diversity
\citep{mishra-etal-2022-cross,chung2024scaling,
wang2023camels}.
Studies such as LIMA and LIMO demonstrate that
carefully selected subsets can achieve strong
performance with limited training data
\citep{zhou2023lima,ye2025limo}.
Existing methods assess sample value through
quality, complexity, diversity, uncertainty,
and semantic similarity
\citep{chen2024alpagasus,lu2024instag,
liu2024deita,cao2024instruction,
li-etal-2024-superfiltering,
kung-etal-2023-active,yang2025ricorefinedincontextcontribution}.
Other studies investigate data pruning,
distributional properties, and sampling strategies
\citep{sorscher2022beyond,maharana2024d2,
nguyen2024qualityvendi,alcoforado-etal-2024-random}.
While these approaches primarily characterize
training examples through sample-level properties,
we investigate how model-internal neuron relevance
can inform data selection.

\subsection{Gradient and Influence-Based Selection}
\label{sec:rw-influence}

Another line of research estimates data value
through parameter gradients and training influence
\citep{grangier-iter-2022-trade,
schioppa2022scaling,bejan-etal-2023-make}.
LESS measures optimization-related similarity
using gradient representations \citep{xia2024less},
while BIDS addresses capability imbalance through
influence normalization
\citep{dai-etal-2025-improving}.
ScaleBiO and LoGra further explore bilevel
optimization and efficient gradient-based
data valuation
\citep{pan-etal-2025-scalebio,choe2025dataworth}.
In contrast to parameter-space influence,
our work examines how attribution-based relevance
is distributed across internal model units.

\subsection{Neuron-Level Signals and Relevance Analysis}
\label{sec:rw-neurons}

Recent studies leverage neuron-level information
for data selection.
NAIT uses reference-derived activation directions
to identify task-relevant examples
\citep{chen2026neuron}, while sparse autoencoder
methods exploit learned features for assessing
task relevance and diversity
\citep{ma2025monosemantic,yang2025sae}.
Other approaches explore activation changes,
neuron relationships, and specialization
\citep{he2026trace,wang2026nag,
tang-etal-2024-language}.

Beyond activation statistics, relevance attribution
offers a complementary perspective on
model-internal signals.
LRP and AttnLRP provide foundations for
relevance-based analysis
\citep{bach2015lrp,achtibat2024attnlrp},
with related applications in neuron functionality
analysis and preference optimization
\citep{le2026crane,
le2026yfpoyokedfeaturepreference}.
However, how different statistical
characterizations of neuron relevance affect
instruction data selection remains underexplored.
We address this question by comparing
\emph{effective relevance support} and
\emph{positive relevance magnitude}
derived from the same relevance profiles.

\section{Method}
\label{sec:method}

We investigate how different characterizations
of neuron relevance affect instruction data selection.
Specifically, we distinguish between the effective
support and magnitude of relevance signals.
Using the same underlying relevance profiles,
we introduce two complementary selection criteria,
ERS and PRM, to examine their selection behaviors
and downstream effectiveness.

\subsection{Problem Formulation}
\label{sec:problem-formulation}

Given a pretrained language model $f_\theta$
and a candidate instruction dataset
$\mathcal{D}=\{d_i\}_{i=1}^{N}$,
our goal is to select a subset
$\mathcal{S}\subseteq\mathcal{D}$ of size $K<N$
for supervised fine-tuning.

Rather than directly estimating each example's
downstream contribution, we construct scoring
functions $s_m(d_i;f_\theta)$ from model-internal
relevance signals, where $m$ denotes the
relevance characterization used for selection.

\subsection{Constructing Neuron Relevance Profiles}
\label{sec:neuron-relevance-profiles}

For an instruction example $d_i$, let
$X_i=(x_1,\ldots,x_{T_i})$ denote its token sequence.
We adopt AttnLRP-guided attribution
\citep{achtibat2024attnlrp}
and define the attribution target as the average
ground-truth next-token logit:

\begin{equation}
\label{eq:attribution-target}
\mathcal{J}_i
=
\frac{1}{T_i-1}
\sum_{t=1}^{T_i-1}
z_\theta(x_{t+1}\mid x_{\leq t}),
\end{equation}

\noindent
where $z_\theta$ denotes the pre-softmax logit.

We extract relevance-related signals from
the output projections of Transformer MLPs.
For layer $\ell$, the output projection is

\begin{equation}
\label{eq:down-projection}
y_{i,t,k}^{(\ell)}
=
\sum_{j=1}^{d_{\mathrm{ffn}}}
W_{kj}^{(\ell)}a_{i,t,j}^{(\ell)},
\end{equation}

\noindent
where
$W_{\mathrm{down}}^{(\ell)}
\in\mathbb{R}^{d_{\mathrm{model}}\times d_{\mathrm{ffn}}}$
and $a_{i,t,j}^{(\ell)}$
denotes the input activation.

Using the backward signals propagated under
AttnLRP-modified rules, we construct a
parameter-level attribution matrix:

\begin{equation}
\label{eq:relevance-matrix}
R_{i,kj}^{(\ell)}
=
\sum_{t=1}^{T_i}
a_{i,t,j}^{(\ell)}
\widetilde{\delta}_{i,t,k}^{(\ell)},
\end{equation}

\noindent
where $\widetilde{\delta}_{i,t,k}^{(\ell)}$
is the modified upstream backward signal.
Aggregating over the projection's input dimension
yields channel-level relevance scores:

\begin{equation}
\label{eq:neuron-relevance}
r_{i,k}^{(\ell)}
=
\sum_{j=1}^{d_{\mathrm{ffn}}}
R_{i,kj}^{(\ell)}.
\end{equation}

These scores represent an operational
AttnLRP-guided relevance measure rather than
standard $\varepsilon$-LRP neuron relevance.

Finally, we concatenate channel-level scores
across layers to obtain the sample-level
\emph{Neuron Relevance Profile}:

\begin{equation}
\label{eq:relevance-profile}
\mathbf{r}_i
=
\operatorname{Concat}_{\ell=1}^{L}
\left(\mathbf{r}_i^{(\ell)}\right).
\end{equation}

\subsection{Relevance Distribution Characterization}
\label{sec:relevance-characterization}

We characterize each relevance profile along
two complementary dimensions:
\emph{effective support} and
\emph{positive magnitude}.

\paragraph{Effective Relevance Support (ERS).}

ERS quantifies how broadly positive relevance
is distributed across channels.
Let $\mathcal{P}_i=\{k\mid r_{i,k}>0\}$
denote the positive relevance indices.
We first normalize positive relevance scores:

\begin{equation}
\label{eq:normalized-relevance}
p_{i,k}
=
\frac{r_{i,k}}
{\sum_{j\in\mathcal{P}_i}r_{i,j}},
\qquad k\in\mathcal{P}_i.
\end{equation}

We then define ERS using the exponential
of Shannon entropy:

\begin{equation}
\label{eq:ers}
\operatorname{ERS}(d_i)
=
\exp\left(
-\sum_{k\in\mathcal{P}_i}
p_{i,k}\log p_{i,k}
\right).
\end{equation}

ERS measures the effective number of channels
supporting positive relevance, accounting for
both their number and distributional uniformity.

\paragraph{Positive Relevance Magnitude (PRM).}

In contrast, PRM measures the average strength
of positive relevance:

\begin{equation}
\label{eq:prm}
\operatorname{PRM}(d_i)
=
\frac{1}{|\mathcal{P}_i|}
\sum_{k\in\mathcal{P}_i}
r_{i,k}.
\end{equation}

When $\mathcal{P}_i=\varnothing$,
both criteria are assigned zero.

Importantly, ERS captures the distributional
spread of relevance and is invariant to uniform
positive rescaling, whereas PRM captures
its absolute magnitude.
Thus, the two criteria characterize distinct
properties of the same relevance profile,
allowing us to investigate how relevance
characterization influences data selection.

\subsection{Relevance-Guided Data Selection}
\label{sec:relevance-guided-selection}

We independently rank candidate examples
using ERS and PRM in descending order.
Given a fixed selection budget $K$,
we select the top-scoring examples:

\begin{equation}
\label{eq:topk-selection}
\mathcal{S}_m
=
\operatorname{TopK}_{d_i\in\mathcal{D}}
s_m(d_i),
\qquad
m\in\{\mathrm{ERS},\mathrm{PRM}\}.
\end{equation}

Both criteria operate on the same candidate
pool and relevance profiles, producing
separate subsets for supervised fine-tuning.
We compare their downstream effectiveness
and analyze their ranking relationships,
selection preferences, and backbone dependence.



\providecommand{\topone}[1]{%
    \textbf{#1}\textsuperscript{\scriptsize 1}}
\providecommand{\toptwo}[1]{%
    #1\textsuperscript{\scriptsize 2}}
\providecommand{\topthree}[1]{%
    #1\textsuperscript{\scriptsize 3}}

\begin{table*}[t]
\centering
\small
\setlength{\tabcolsep}{5pt}
\renewcommand{\arraystretch}{1.08}

\begin{tabular}{llccccc@{\hspace{8pt}}c}
\toprule

\textbf{Backbone}
& \textbf{Method}
& \textbf{ARC-C}
& \textbf{CSQA}
& \textbf{Meta-GSM8K}
& \textbf{IFEval}
& \textbf{WinoGrande}
& \textbf{Avg. Rank} $\downarrow$ \\

\midrule


\multirow{7}{*}{Llama-3.2-1B}
& \textit{Base}
& 39.68 & 52.01 & 6.75 & 10.54 & 60.30 & -- \\

\cmidrule(lr){2-8}

& Random
& 40.87
& 49.63
& \topone{18.14}
& \topone{16.64}
& \topone{63.77}
& \toptwo{2.40} \\

& Activation
& 40.44
& \topone{50.86}
& 13.08
& \toptwo{16.45}
& \topthree{63.22}
& 3.40 \\

& Gradient
& \toptwo{41.72}
& 49.55
& 13.08
& \topone{16.64}
& 62.75
& 3.40 \\

& NAIT
& 41.21
& \topthree{50.20}
& \topthree{15.19}
& 11.83
& \topone{63.77}
& \topthree{3.00} \\

& \textbf{PRM}
& \topone{42.24}
& \toptwo{50.61}
& \toptwo{16.46}
& \topthree{13.86}
& \topthree{63.22}
& \topone{2.20} \\

& \textbf{ERS}
& \topthree{41.38}
& \topone{50.86}
& 14.35
& \topone{16.64}
& \toptwo{63.69}
& \topone{2.20} \\

\midrule


\multirow{7}{*}{Qwen2-0.5B}
& \textit{Base}
& 31.06 & 50.78 & 36.29 & 13.12 & 58.09 & -- \\

\cmidrule(lr){2-8}

& Random
& \toptwo{32.17}
& \topone{50.86}
& 37.97
& \toptwo{15.71}
& \toptwo{56.51}
& \toptwo{2.20} \\

& Activation
& 31.57
& 49.63
& \topthree{39.66}
& 14.23
& \topthree{56.43}
& 4.00 \\

& Gradient
& \topthree{31.91}
& \topthree{50.45}
& 32.49
& \topone{16.27}
& 55.88
& \topthree{3.40} \\

& NAIT
& 31.31
& 50.29
& 35.86
& 12.94
& \topone{57.06}
& 4.20 \\

& \textbf{PRM}
& 31.57
& 50.12
& \topone{41.77}
& 14.05
& \topthree{56.43}
& 3.60 \\

& \textbf{ERS}
& \topone{32.59}
& \toptwo{50.70}
& \toptwo{40.08}
& \topthree{15.53}
& \topone{57.06}
& \topone{1.80} \\

\midrule


\multirow{7}{*}{Qwen2-1.5B}
& \textit{Base}
& 44.45 & 67.40 & 59.49 & 16.08 & 66.38 & -- \\

\cmidrule(lr){2-8}

& Random
& 43.17
& \toptwo{69.53}
& \toptwo{60.76}
& 19.59
& 65.67
& 3.60 \\

& Activation
& 43.34
& 69.21
& 58.23
& \topthree{20.15}
& \topone{66.38}
& 3.40 \\

& Gradient
& \topone{44.54}
& \topone{69.78}
& 56.96
& 17.56
& \topone{66.38}
& \toptwo{2.80} \\

& NAIT
& 43.26
& \topthree{69.45}
& 59.07
& \topone{23.11}
& \toptwo{66.14}
& \topthree{3.00} \\

& \textbf{PRM}
& \toptwo{43.60}
& \toptwo{69.53}
& \topthree{60.34}
& \topthree{20.15}
& \toptwo{66.14}
& \topone{2.40} \\

& \textbf{ERS}
& \topthree{43.43}
& 68.71
& \topone{61.60}
& \toptwo{21.44}
& \topthree{65.98}
& \toptwo{2.80} \\

\bottomrule
\end{tabular}

\caption{
Maximum downstream performance (\%) across three
training-epoch checkpoints for each benchmark.
Superscripts $1$, $2$, and $3$ denote the top three
selection methods, with the best highlighted in bold.
Avg. Rank is averaged across five benchmarks
(lower is better).
\textit{Base} is excluded from ranking.
}

\label{tab:main_results}
\end{table*}


\begin{table}[t]
\centering
\small
\setlength{\tabcolsep}{2.8pt}
\renewcommand{\arraystretch}{1.08}

\begin{tabular}{lccccc}
\toprule

\textbf{Method}
& \textbf{Qwen0.5}
& \textbf{Qwen1.5}
& \textbf{Llama1B}
& \shortstack{\textbf{Overall}\\\textbf{Max} $\downarrow$}
& \shortstack{\textbf{Overall}\\\textbf{Mean} $\downarrow$} \\

\midrule

Random
& 2.20 & 3.60 & 2.40 & 2.73 & 2.87 \\

Activation
& 4.00 & 3.40 & 3.40 & 3.60 & 3.73 \\

Gradient
& 3.40 & 2.80 & 3.40 & 3.20 & 4.00 \\

NAIT
& 4.20 & 3.00 & 3.00 & 3.40 & 3.93 \\

\midrule

\textbf{PRM}
& 3.60 & \textbf{2.40} & \textbf{2.20}
& 2.73 & 3.07 \\

\textbf{ERS}
& \textbf{1.80} & 2.80 & \textbf{2.20}
& \textbf{2.27} & \textbf{2.73} \\

\bottomrule
\end{tabular}

\caption{
Average ranks across five downstream benchmarks
(lower is better).
For each backbone and benchmark, the six selection
methods are ranked by either their maximum (Max)
or mean (Mean) performance over three training epochs.
Backbone-specific columns report Max-based
average ranks.
Overall Max and Overall Mean further average
the corresponding backbone-specific ranks
across all three models.
Bold indicates the best result in each column,
including ties.
}

\label{tab:aggregate_ranks}
\end{table}

\begin{figure}[t]
\centering
\includegraphics[
    width=\columnwidth
]{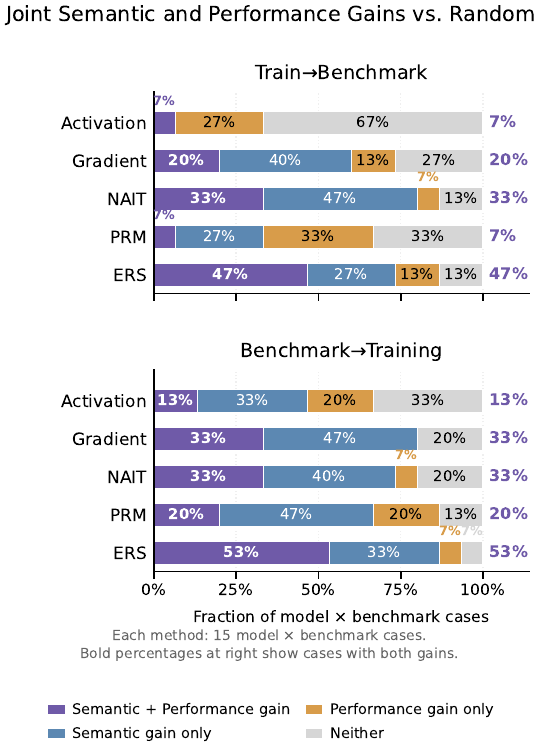}

\caption{
Joint semantic proximity and downstream performance
changes relative to Random across 15
backbone--benchmark combinations.
Purple denotes improvements in both metrics;
blue and orange indicate improvements only in semantic
proximity or downstream performance, respectively;
gray indicates neither.
Downstream performance is measured using Max scores.
}
\label{fig:semantic_performance_quadrants}
\end{figure}

\begin{figure*}[t]
\centering
\includegraphics[
    width=0.9\textwidth
]{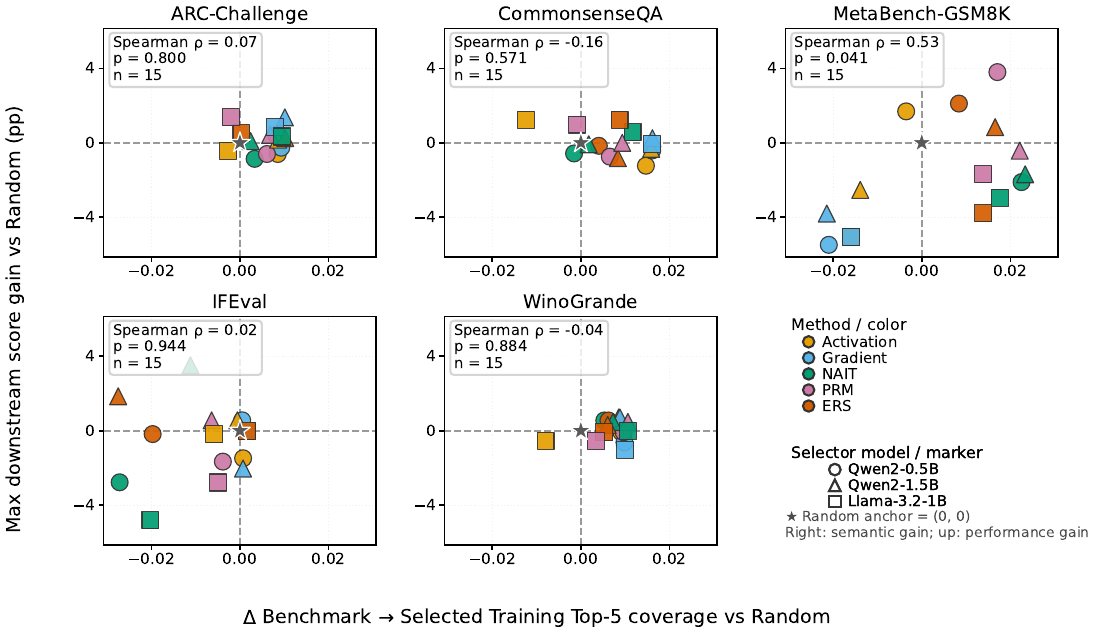}

\caption{
Relationship between changes in
Benchmark-to-Training Top-5 semantic proximity
and downstream performance gains relative to Random.
Each benchmark contains 15 observations.
Colors indicate selection methods and markers
indicate scoring backbones.
Performance gains are measured in percentage points
using Max scores.
Spearman rank correlations are reported for each
benchmark.
}
\label{fig:semantic_performance_correlation}
\end{figure*}

\section{Experiments}
\label{sec:experiments}

\subsection{Experimental Setup}
\label{sec:experimental_setup}

\paragraph{Datasets and Models.}
We construct a candidate pool of approximately 100K
instruction examples from Tülu 3
\citep{lambert2024tulu3}\footnote{
\url{https://huggingface.co/datasets/allenai/tulu-3-sft-mixture}}
and select 30K examples under a fixed data budget.
We evaluate three pretrained models:
Qwen2-0.5B and Qwen2-1.5B \citep{yang2024qwen2},
and Llama-3.2-1B \citep{grattafiori2024llama3}.

\paragraph{Baselines.}
We compare six selection methods:
Random, Activation, Gradient, NAIT,
ERS, and PRM.

\textbf{Random} uniformly samples examples.
\textbf{Activation} adapts FFN activation
probability from LAPE
\citep{tang-etal-2024-language}.
\textbf{Gradient} uses gradient-times-activation
neuron attribution
\citep{leng-xiong-2025-towards}.
\textbf{NAIT} \citep{chen2026neuron}
uses reference-derived representation directions.

Unlike NAIT, ERS and PRM require no
target-task reference examples.
All methods share the same candidate pool
and selection budget.
Implementation details are provided in
Appendix~\ref{sec:appendix_baselines}.

\paragraph{Implementation Details.}
We perform full-parameter SFT using LLaMA-Factory
with AdamW and cosine learning rate scheduling.
All models are trained for three epochs
with sequence packing.
Training configurations and relevance computation
details are provided in
Appendix~\ref{sec:appendix_training_details}.

\paragraph{Task Categorization.}
We categorize instruction examples into nine
capability categories based on their source datasets
to analyze task-specific selection preferences.
Category definitions and candidate pool distributions
are provided in
Appendix~\ref{sec:appendix_task_distribution}.

\paragraph{Evaluation.}
We evaluate on five benchmarks:
ARC-Challenge (ARC-C; 25-shot)
\citep{clark2018think},
CommonsenseQA (CSQA; 0-shot)
\begin{NoHyper}
\citep{talmor-etal-2019-commonsenseqa}
\end{NoHyper},
MetaBench-GSM8K (fixed 5-shot)
\citep{kipnis2025metabench,cobbe2021gsm8k},
IFEval (0-shot) \citep{zhou2023instruction},
and WinoGrande (0-shot)
\citep{DBLP:journals/corr/abs-1907-10641}.

We report normalized accuracy for ARC-C,
accuracy for CSQA and WinoGrande,
exact match for MetaBench-GSM8K,
and prompt-level strict accuracy for IFEval.
We evaluate all three epoch checkpoints,
reporting their mean (\textit{Mean}) and
maximum (\textit{Max}) performance.
We also report average performance ranks
across backbones and benchmarks,
where lower is better.

\subsection{Main Results}
\label{sec:main_results}

\paragraph{Overall Effectiveness.}
Table~\ref{tab:main_results} reports the
maximum downstream performance across three
training epochs, with mean results provided in
Appendix Table~\ref{tab:mean_main_results}.
As shown in Table~\ref{tab:aggregate_ranks},
\textbf{ERS achieves the best overall average
rank under both Max (2.27) and Mean (2.73)
evaluation} among the six selection methods.
These results demonstrate the practical
effectiveness of relevance support as a
data selection criterion.

\paragraph{Support versus Magnitude.}
Although ERS and PRM are derived from the
same relevance profiles, their downstream
effectiveness varies across model backbones.
ERS achieves the best average rank of 1.80
on Qwen2-0.5B and ties with PRM at 2.20
on Llama-3.2-1B.
In contrast, PRM performs best on Qwen2-1.5B
with a rank of 2.40, compared with 2.80 for ERS.
At the benchmark level, both criteria achieve
leading results on different tasks.
For instance, ERS obtains the highest
MetaBench-GSM8K score of 61.60 on Qwen2-1.5B,
while PRM achieves the highest ARC-C score
of 42.24 on Llama-3.2-1B.

These findings suggest that
\textbf{effective support provides strong
aggregate performance, while the relative
utility of support and magnitude depends
on the model backbone and downstream task.}

\paragraph{Semantic Proximity and Downstream Performance.}
We further examine whether semantic proximity
to downstream benchmarks explains selection
effectiveness.
As shown in
Figure~\ref{fig:semantic_performance_quadrants},
ERS achieves the highest proportion of joint
semantic proximity and performance improvements
over Random (47\% and 53\% across two directions).
However,
Figure~\ref{fig:semantic_performance_correlation}
reveals that correlations between semantic proximity
and performance vary across benchmarks and are
generally weak.
These findings suggest that semantic proximity
alone cannot fully explain downstream performance,
motivating further analysis of relevance-based
selection behaviors.
Detailed methodology and results are provided in
Appendix~\ref{sec:appendix_semantic_quadrants}.

\begin{table}[t]
\centering
\normalsize
\setlength{\tabcolsep}{4pt}
\begin{tabular}{@{}lrrr@{}}
\toprule
\textbf{Signal Pair}
& \textbf{Raw $\rho$}
& \textbf{Strat. $\rho$}
& \textbf{Overlap (\%)} \\
\midrule

\multicolumn{4}{@{}l}{\textbf{Qwen2-0.5B}} \\
ERS / Activation & -0.169 &  0.123 & 22.05 \\
ERS / Gradient   & -0.402 & -0.032 & 18.58 \\
ERS / PRM        & -0.088 & -0.009 & 25.41 \\
PRM / Activation &  0.612 &  0.412 & 72.17 \\
PRM / Gradient   &  0.440 &  0.140 & 60.57 \\

\midrule
\multicolumn{4}{@{}l}{\textbf{Qwen2-1.5B}} \\
ERS / Activation & -0.345 & -0.138 & 18.34 \\
ERS / Gradient   & -0.277 & -0.044 & 23.85 \\
ERS / PRM        &  0.137 &  0.232 & 36.74 \\
PRM / Activation &  0.590 &  0.446 & 66.14 \\
PRM / Gradient   &  0.355 &  0.130 & 56.20 \\

\midrule
\multicolumn{4}{@{}l}{\textbf{Llama-3.2-1B}} \\
ERS / Activation & -0.042 & -0.076 & 32.56 \\
ERS / Gradient   &  0.204 &  0.039 & 41.67 \\
ERS / PRM        &  0.485 &  0.515 & 56.78 \\
PRM / Activation &  0.146 &  0.119 & 43.09 \\
PRM / Gradient   &  0.100 &  0.048 & 38.69 \\

\bottomrule
\end{tabular}
\caption{
Ranking correlations and selection overlap between
relevance-based and baseline methods.
Raw $\rho$ denotes Spearman rank correlation over
jointly valid examples.
Stratified $\rho$ is the residual rank correlation
after centering global ranks within joint strata
defined by task type, answer length, and answer
perplexity.
Top-$K$ overlap is
$|S_A\cap S_B|/\min(|S_A|,|S_B|)$,
computed over deduplicated valid selections.
}
\label{tab:signal_correlation}
\end{table}

\begin{figure}[t]
\centering
\includegraphics[
    width=0.5\textwidth
]{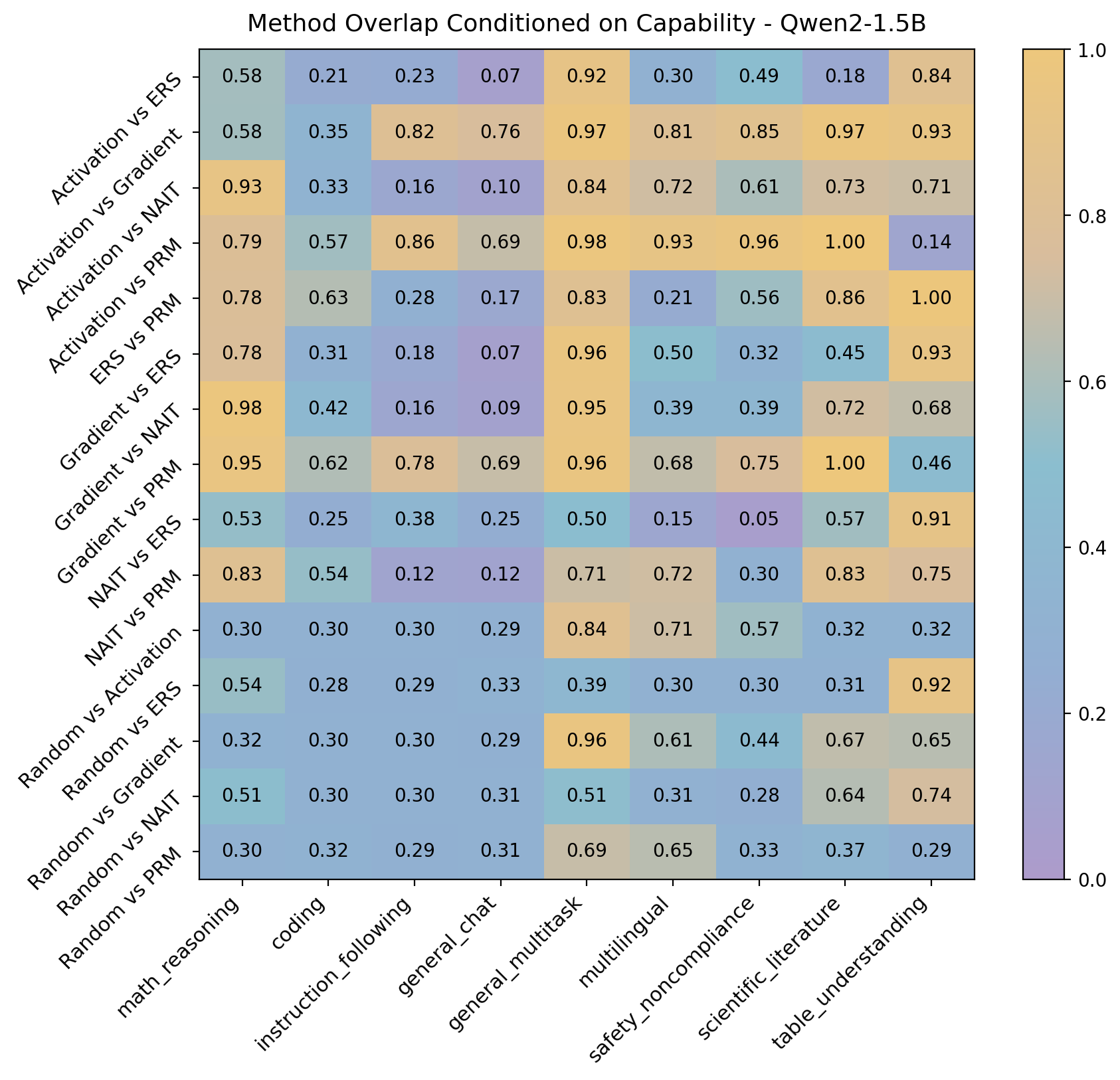}
\caption{
Task-conditioned selection overlap among
different methods on Qwen2-1.5B.
Each cell reports the overlap between two
methods within a task category, normalized
by the smaller selected subset.
Cells with fewer than 50 selected examples
in either subset are masked.
Results for Qwen2-0.5B and Llama-3.2-1B
are provided in
Appendix~\ref{sec:appendix_task_overlap}.
}
\label{fig:task_conditioned_overlap}
\end{figure}

\begin{figure*}[t]
\centering
\includegraphics[
    width=0.98\textwidth
]{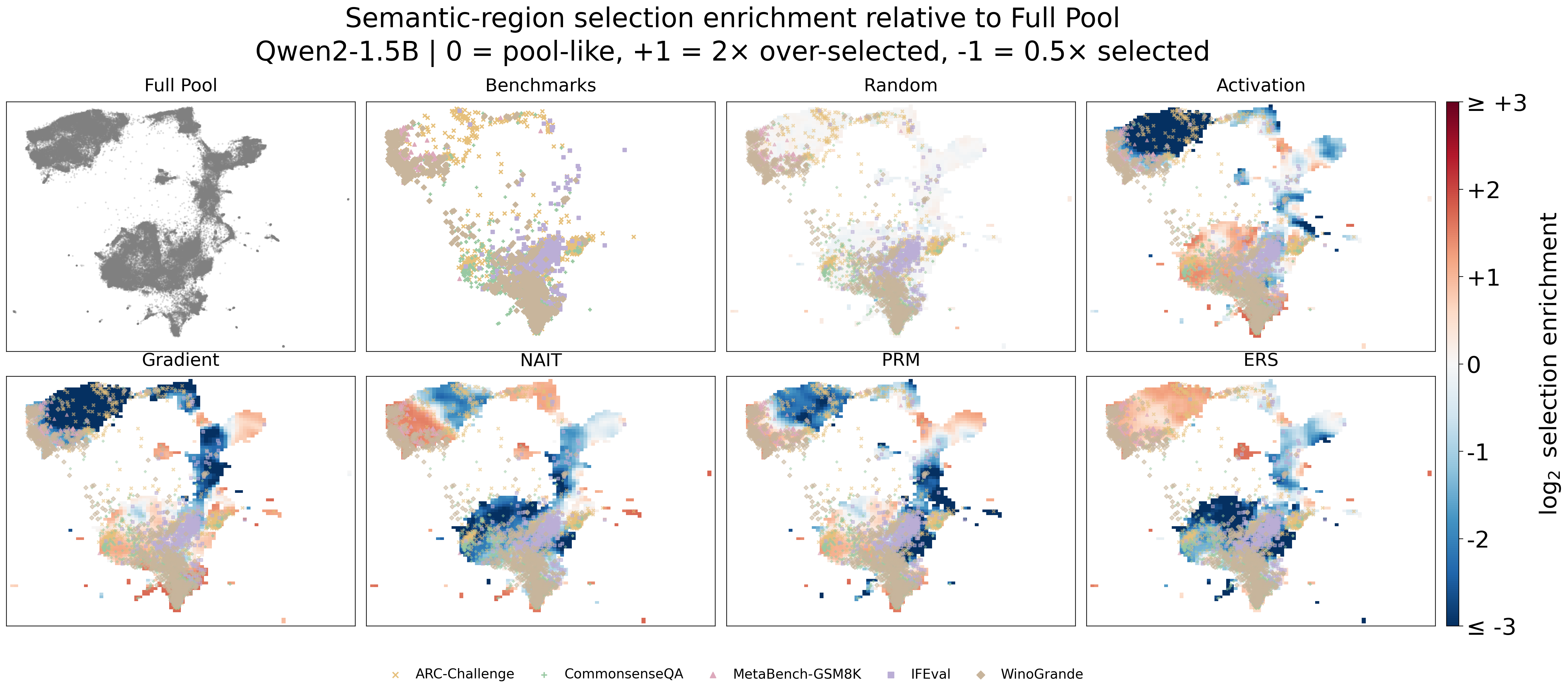}

\caption{
Semantic-region selection enrichment on Qwen2-1.5B.
All methods share the same UMAP projection.
Red and blue indicate over-selection and
under-selection relative to the full candidate
pool, respectively.
Markers denote examples from the five downstream
benchmarks.
Corresponding results for Qwen2-0.5B and
Llama-3.2-1B are provided in
Appendix~\ref{sec:appendix_umap}.
}
\label{fig:selection_umap}

\end{figure*}

\begin{table}[t]
\centering
\normalsize
\setlength{\tabcolsep}{8pt}
\begin{tabular}{lccc}
\toprule
\textbf{Method}
& \textbf{Q0.5/Q1.5}
& \textbf{Q0.5/L1}
& \textbf{Q1.5/L1} \\
\midrule
Random
& 100.00\% & 100.00\% & 100.00\% \\
Activation
& 89.49\% & 58.99\% & 55.92\% \\
Gradient
& 93.00\% & 87.77\% & 86.78\% \\
NAIT
& 53.25\% & 54.05\% & 39.74\% \\
PRM
& 79.91\% & 51.90\% & 54.43\% \\
\textbf{ERS}
& \textbf{53.20\%}
& \textbf{30.74\%}
& \textbf{34.66\%} \\
\bottomrule
\end{tabular}
\caption{
Cross-backbone selection overlap.
Q0.5, Q1.5, and L1 denote Qwen2-0.5B,
Qwen2-1.5B, and Llama-3.2-1B.
Overlap is computed over deduplicated valid
selections after excluding the
\texttt{assistant\_identity} category, using
$|S_A\cap S_B|/\min(|S_A|,|S_B|)$.
Random uses the same random selection across
backbones.
}
\label{tab:cross_backbone_overlap}
\end{table}

\section{Analysis}
\label{sec:analysis}

We investigate how relevance characterization
affects instruction data selection from three
perspectives: selection preferences,
relationships with existing signals,
and backbone dependence.


\subsection{Effects of Relevance Characterization}
\label{sec:relevance_characterization}

ERS and PRM characterize the effective support
and average magnitude of the same relevance
profiles, respectively.
We examine whether these different
characterizations lead to distinct
selection behaviors.

\paragraph{Ranking and Selection Agreement.}
As shown in Table~\ref{tab:signal_correlation},
ERS and PRM exhibit weak ranking correlations
on Qwen2-0.5B ($\rho=-0.088$) and
Qwen2-1.5B ($\rho=0.137$), compared with
a moderate correlation on Llama-3.2-1B
($\rho=0.485$).
Their Top-$K$ overlaps are 25.41\%,
36.74\%, and 56.78\%, respectively.

These results indicate that
\textbf{different characterizations of the same
relevance profile can produce distinct
sample rankings and selection preferences},
with varying degrees of agreement across backbones.

\paragraph{Task-Conditioned Selection.}
We further examine whether these differences
persist within individual task categories.
Figure~\ref{fig:task_conditioned_overlap}
shows substantial variation in selection
agreement across tasks.
On Qwen2-1.5B, ERS and PRM achieve overlaps
of 0.17 and 0.21 on General Chat and
Multilingual examples, respectively,
compared with 0.83 on General Multitask.

These findings suggest that
\textbf{selection differences are not solely
explained by overall task composition}.
Different relevance characteristics can
favor different examples even within the
same task category.
Additional capability compositions are
provided in
Appendix~\ref{sec:appendix_capability_composition}.

\paragraph{Semantic Selection Patterns.}
We further compare selection preferences
in a shared semantic embedding space.
Using a common UMAP projection, we measure
local selection enrichment relative to
the full candidate pool.
As shown in Figure~\ref{fig:selection_umap},
ERS and PRM exhibit different patterns
of semantic enrichment and under-selection.
Task-type enrichment further reveals that
ERS emphasizes Math Reasoning (1.82$\times$)
and Table Understanding (3.05$\times$),
whereas PRM favors General Multitask
(2.29$\times$) and Multilingual
(2.13$\times$) examples
(Appendix Table~\ref{tab:task_enrichment}).

Together, these analyses show that
\textbf{relevance characterization influences
not only sample rankings, but also
within-task preferences, task composition,
and semantic selection patterns}.


\subsection{Relationship with Existing Selection Signals}
\label{sec:signal_relationship}

We examine how ERS and PRM relate to
activation- and gradient-based selection.
We compute Spearman rank correlations
and additionally center ranks within
joint strata defined by task type,
answer length, and perplexity. Table~\ref{tab:signal_correlation} reveals
different relationships with existing signals.
On Qwen2-1.5B, ERS exhibits weak or negative
correlations with Activation ($\rho=-0.345$)
and Gradient ($\rho=-0.277$),
whereas PRM correlates more strongly
with Activation ($\rho=0.590$).

After stratification, the correlations
between ERS and Activation or Gradient
remain weak ($-0.138$ and $-0.044$),
indicating that these differences are
not fully accounted for by the
considered data characteristics.
Similar contrasting relationships are
observed on Qwen2-0.5B, while their
strengths differ on Llama-3.2-1B.
These findings suggest that
\textbf{effective relevance support captures
ranking information less aligned with
conventional activation-based selection
than relevance magnitude}, particularly
on Qwen2 backbones. Additional answer NLL analyses are provided
in Appendix~\ref{sec:appendix_nll_distribution}.


\subsection{Backbone Dependence}
\label{sec:backbone_dependence}

Finally, we examine how scoring backbones
affect the selected subsets.
Table~\ref{tab:cross_backbone_overlap}
shows that ERS achieves a cross-backbone
overlap of 53.20\% between Qwen2-0.5B
and Qwen2-1.5B, compared with 79.91\%
for PRM.
For comparisons involving Llama-3.2-1B,
ERS overlaps range from 30.74\% to 34.66\%,
whereas PRM achieves 51.90\% to 54.43\%.
These results indicate that
\textbf{ERS exhibits stronger backbone
dependence than PRM} in the evaluated models.
This variation is accompanied by differences
in downstream effectiveness:
ERS ranks first on Qwen2-0.5B and ties
with PRM on Llama-3.2-1B, while PRM
performs best on Qwen2-1.5B.

Overall, our analyses demonstrate that
\textbf{the characterization of neuron relevance
shapes data selection behavior},
and that these effects vary with
the scoring backbone.

\section{Conclusion}
\label{sec:conclusion}

In this work, we investigate how different
characterizations of neuron relevance affect
instruction data selection.
Using AttnLRP-guided relevance profiles,
we introduce Effective Relevance Support (ERS)
and Positive Relevance Magnitude (PRM) to
characterize the effective support and average
strength of positive relevance, respectively.
Our analyses reveal that, despite sharing
the same relevance profiles, ERS and PRM
produce distinct sample rankings, task-conditioned
preferences, and semantic selection patterns.
On Qwen2 models, PRM aligns more closely
with activation-based selection, whereas ERS
exhibits substantially different ranking behavior.
These differences also vary across model backbones.
Experiments across three language models and
five downstream benchmarks show that ERS achieves
the best overall average performance rank
under both Mean and Max evaluation, although
the relative effectiveness of the two criteria
varies across backbones.
Overall, our findings highlight that
\textbf{how neuron relevance is distributed
and characterized, rather than its magnitude
alone, matters for instruction data selection.}

\section*{Limitations}

Our experiments focus on three relatively small
language models with a fixed candidate pool and
selection budget.
Future work will investigate the generalizability
of relevance-based selection to larger models,
different data distributions, and varying
selection budgets, as well as its robustness
across training runs.


\section*{Ethical Considerations}

This work uses existing instruction datasets and language
models without collecting new personal data.
Relevance-guided data selection may alter the task and
language distributions of training subsets, potentially
affecting capability coverage and model biases.
We do not systematically evaluate its implications
for safety or fairness.

\bibliography{custom}

\appendix

\appendix
\label{sec:appendix}


\section{Additional Selection Analyses}
\label{sec:appendix_selection}

\subsection{Task-Type Enrichment}
\label{sec:appendix_task_enrichment}

We analyze the relative representation of each
task category in the selected subsets.
For a selection method $m$ and task category $c$,
we define the enrichment ratio as

\begin{equation}
E_m(c)
=
\frac{P(c\mid S_m)}
{P(c\mid \mathcal{D})},
\end{equation}

where $S_m$ denotes the selected subset and
$\mathcal{D}$ denotes the full candidate pool.
A value above one indicates over-representation
relative to the candidate pool.

\section{Semantic Proximity and Performance Consistency}
\label{sec:appendix_semantic_quadrants}

\paragraph{Semantic Proximity.}
We measure the semantic proximity between
selected training examples and downstream
benchmarks using BGE-M3 embeddings.
For training examples, we embed user queries
without assistant responses.
For benchmarks, we embed questions and
available answer options, excluding gold answers.

We compute cosine similarity between training
and benchmark embeddings and consider two
directions: Training-to-Benchmark (T$\rightarrow$B)
and Benchmark-to-Training (B$\rightarrow$T).
Both directions use the median Top-5 mean
cosine similarity as the primary metric.

Given a selected training set $S$ and a benchmark
set $B$, the Training-to-Benchmark proximity is

\begin{equation}
\label{eq:train_to_benchmark_proximity}
P_{T\rightarrow B}(S,B)
=
\underset{x\in S}{\operatorname{median}}
\left[
\frac{1}{5}
\sum_{y\in\mathcal{N}_5(x,B)}
\operatorname{sim}(x,y)
\right],
\end{equation}

where $\mathcal{N}_5(x,B)$ denotes the five
nearest benchmark examples to $x$ under
cosine similarity.

Similarly, Benchmark-to-Training proximity
is defined as

\begin{equation}
\label{eq:benchmark_to_train_proximity}
P_{B\rightarrow T}(S,B)
=
\underset{y\in B}{\operatorname{median}}
\left[
\frac{1}{5}
\sum_{x\in\mathcal{N}_5(y,S)}
\operatorname{sim}(y,x)
\right].
\end{equation}

The former characterizes the semantic proximity
of selected training examples to a benchmark,
while the latter measures how well benchmark
examples are covered by the selected training set.

\paragraph{Relative Semantic and Performance Gains.}
We use Random selection as the reference
for both semantic proximity and downstream
performance.

For selection method $m$, scoring backbone $b$,
benchmark $t$, and semantic direction $d$,
the semantic proximity gain is

\begin{equation}
\label{eq:semantic_gain}
\Delta_{\mathrm{sem}}^{d}(m,b,t)
=
P_d(S_{m,b},B_t)
-
P_d(S_{\mathrm{Random},b},B_t),
\end{equation}

where $S_{m,b}$ denotes the training subset
selected by method $m$ using backbone $b$,
and $B_t$ denotes the benchmark examples.

The downstream performance gain is defined as

\begin{equation}
\label{eq:performance_gain}
\Delta_{\mathrm{perf}}(m,b,t)
=
Q_{m,b,t}^{\mathrm{Max}}
-
Q_{\mathrm{Random},b,t}^{\mathrm{Max}},
\end{equation}

where $Q^{\mathrm{Max}}$ is the maximum
benchmark score across the three training-epoch
checkpoints.
Performance differences are reported in
percentage points.

\paragraph{Joint Improvement Classification.}
For each backbone--benchmark pair, we classify
the results into four mutually exclusive
categories based on semantic proximity and
downstream performance gains:
\textit{Both Gain}
($\Delta_{\mathrm{sem}}>0$,
$\Delta_{\mathrm{perf}}>0$),
\textit{Semantic Only}
($\Delta_{\mathrm{sem}}>0$,
$\Delta_{\mathrm{perf}}\leq0$),
\textit{Performance Only}
($\Delta_{\mathrm{sem}}\leq0$,
$\Delta_{\mathrm{perf}}>0$),
and \textit{Neither}
($\Delta_{\mathrm{sem}}\leq0$,
$\Delta_{\mathrm{perf}}\leq0$).
Zero differences are treated as no improvement.

\paragraph{Aggregation Across Backbones and Benchmarks.}
We evaluate three scoring backbones and five
downstream benchmarks, yielding 15
backbone--benchmark combinations per method.

For semantic direction $d$, the proportion
of combinations exhibiting simultaneous
semantic and performance improvements is

\begin{equation}
\label{eq:joint_improvement_ratio}
\begin{aligned}
J_m^d
&=
\frac{1}{15}
\sum_{b=1}^{3}
\sum_{t=1}^{5}
\mathbb{I}\bigl[
\Delta_{\mathrm{sem}}^d(m,b,t)>0
\\
&\qquad\land\,
\Delta_{\mathrm{perf}}(m,b,t)>0
\bigr].
\end{aligned}
\end{equation}

where $\mathbb{I}[\cdot]$ denotes the
indicator function.

We compute the proportions of all four
categories analogously.
Figure~\ref{fig:semantic_performance_quadrants}
visualizes these proportions as stacked bars
for each selection method and semantic direction.

This analysis quantifies the frequency of
joint improvements rather than establishing
a causal relationship between semantic
proximity and downstream performance.
As shown in
Figure~\ref{fig:semantic_performance_quadrants},
ERS achieves the highest proportion of joint
semantic proximity and performance improvements,
reaching 47\% and 53\% in the two directions,
respectively.

\paragraph{Semantic Proximity--Performance Correlation.}
We further examine the relationship between
semantic proximity changes and downstream
performance gains using Spearman rank correlation,
as shown in Figure~\ref{fig:semantic_performance_correlation}.

For each backbone $b$, benchmark $t$, and
non-random selection method $m$, we construct
a paired observation $(x_{m,b,t}, y_{m,b,t})$,
where

\begin{equation}
\begin{aligned}
x_{m,b,t}
&= \Delta_{\mathrm{sem}}^{B\rightarrow T}(m,b,t),\\
y_{m,b,t}
&= \Delta_{\mathrm{perf}}(m,b,t).
\end{aligned}
\end{equation}

Here, $x_{m,b,t}$ denotes the change in
Benchmark-to-Training Top-5 semantic proximity
relative to Random, and $y_{m,b,t}$ denotes
the corresponding downstream performance gain
in percentage points using Max scores.

For each benchmark, we aggregate observations
from three scoring backbones and five non-random
selection methods, yielding 15 paired observations.
We then compute the Spearman rank correlation:

\begin{equation}
\rho_t =
\operatorname{corr}
\left(
\operatorname{rank}(\mathbf{x}_t),
\operatorname{rank}(\mathbf{y}_t)
\right),
\end{equation}

where $\mathbf{x}_t$ and $\mathbf{y}_t$
contain the semantic proximity changes and
performance gains for benchmark $t$,
respectively.
Tied values receive average ranks.

We report two-sided $p$-values obtained using
SciPy's Spearman correlation test.
The Random baseline serves as the origin
$(0,0)$ for visualization but is excluded
from correlation computation.
No multiple-testing correction is applied
across benchmarks.
As shown in
Figure~\ref{fig:semantic_performance_correlation},
MetaBench-GSM8K exhibits a positive correlation
($\rho=0.53$), whereas the remaining benchmarks
show weak correlations.
These results suggest that semantic proximity
does not consistently predict downstream
performance improvements.


\section{Task Categorization and Candidate Pool Distribution}
\label{sec:appendix_task_distribution}

\paragraph{Task Categorization.}

We categorize instruction examples according
to their source datasets in the T\"ulu-3
candidate pool.
Specifically, each dataset source is assigned
to one of nine predefined capability categories:
\textit{Math Reasoning},
\textit{Coding},
\textit{Instruction Following},
\textit{General Chat},
\textit{General Multitask},
\textit{Multilingual},
\textit{Safety and Non-compliance},
\textit{Scientific Literature},
and \textit{Table Understanding}.

This source-based categorization provides
a consistent task taxonomy for analyzing
the composition and selection preferences
of different data selection methods.

Examples belonging to the
\texttt{assistant\_identity} category are
excluded from the task-level distribution
and overlap analyses.

\paragraph{Candidate Pool Distribution.}

\begin{figure*}[t]
\centering
\includegraphics[
    width=0.75\textwidth
]{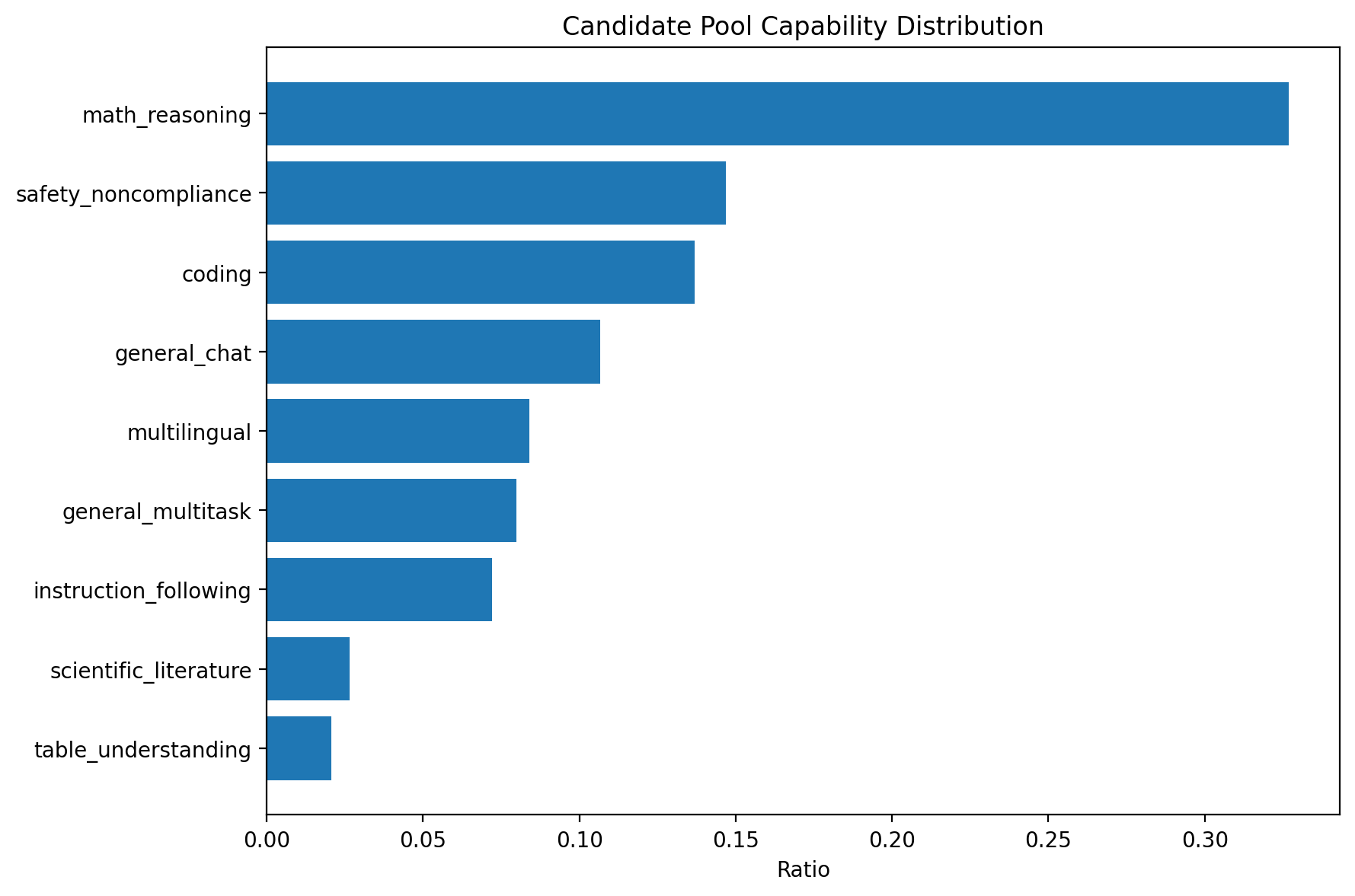}

\caption{
Distribution of capability categories in
the candidate training pool.
Instruction examples are categorized
according to their source datasets.
Proportions are calculated after excluding
the \texttt{assistant\_identity} category
from task-level analysis.
}
\label{fig:candidate_pool_task_distribution}
\end{figure*}

Figure~\ref{fig:candidate_pool_task_distribution}
presents the distribution of capability
categories in the candidate training pool.

The proportion of each category is computed
as the number of examples assigned to that
category divided by the total number of
examples retained for task-level analysis.

The candidate pool exhibits an imbalanced
task distribution, with Math Reasoning
representing the largest category, followed
by Safety and Non-compliance and Coding.
These proportions provide a reference
for interpreting task-specific selection
preferences and enrichment patterns.


\section{Additional Task-Conditioned Selection Overlap}
\label{sec:appendix_task_overlap}

We provide additional task-conditioned
selection overlap results for Qwen2-0.5B
and Llama-3.2-1B, complementing the
Qwen2-1.5B results presented in
Figure~\ref{fig:task_conditioned_overlap}.

For each task category, overlap is computed
between the selected subsets of two methods
and normalized by the smaller subset size.
Comparisons involving fewer than 50 selected
examples in either subset are masked.
All heatmaps use the same color scale
ranging from 0 to 1.

\section{Additional Semantic Selection Patterns}
\label{sec:appendix_umap}

We provide additional semantic-region selection
enrichment visualizations for Qwen2-0.5B and
Llama-3.2-1B, complementing the Qwen2-1.5B
results shown in Figure~\ref{fig:selection_umap}.
All visualizations use the same UMAP projection
and color scale to facilitate comparison
across scoring backbones.
\paragraph{Semantic-region Enrichment.}

To characterize the semantic preferences of
different data selection methods, we compute
local selection enrichment in a shared
two-dimensional UMAP space.

We partition the UMAP projection into a
$90 \times 90$ grid.
For each region $r$, we count the number
of candidate pool examples and selected
training examples falling into that region.

We apply Gaussian smoothing with
$\sigma=1.0$ to both count distributions
and normalize them independently.

Let $P_m(r)$ denote the normalized
smoothed density of examples selected by
method $m$ in region $r$, and let
$P_{\mathcal{D}}(r)$ denote the corresponding
density of the full candidate pool.

The semantic-region enrichment is defined as

\begin{equation}
E_m(r)
=
\log_2
\frac{P_m(r)}
{P_{\mathcal{D}}(r)}.
\label{eq:semantic_region_enrichment}
\end{equation}

Positive values indicate that a semantic
region is over-represented in the selected
subset relative to the candidate pool,
whereas negative values indicate
under-representation.

Specifically, $E_m(r)=0$ indicates equal
relative density, while values of $+1$
and $-1$ correspond to twofold and
half-density representation, respectively.

To avoid unreliable estimates in sparse
regions, we mask grid cells containing
fewer than three original candidate pool
examples.

For visualization, enrichment values are
displayed using a fixed symmetric color
range of $[-3,3]$ across all methods and
backbones.
Values outside this range are color-saturated
without modifying the underlying estimates.


\section{Additional Analysis of Capability Composition}
\label{sec:appendix_capability_composition}

We further examine the capability composition
of training subsets selected by different
data selection methods across three backbones:
Qwen2-0.5B, Qwen2-1.5B, and Llama-3.2-1B.

Following the source-based task categorization
described in
Appendix~\ref{sec:appendix_task_distribution},
we group selected training examples into nine
capability categories.

For each method, we compute the proportion
of selected examples belonging to each
category, excluding the
\texttt{assistant\_identity} category
from the task-level analysis.

\begin{figure*}[t]
\centering
\includegraphics[
    width=0.95\textwidth
]{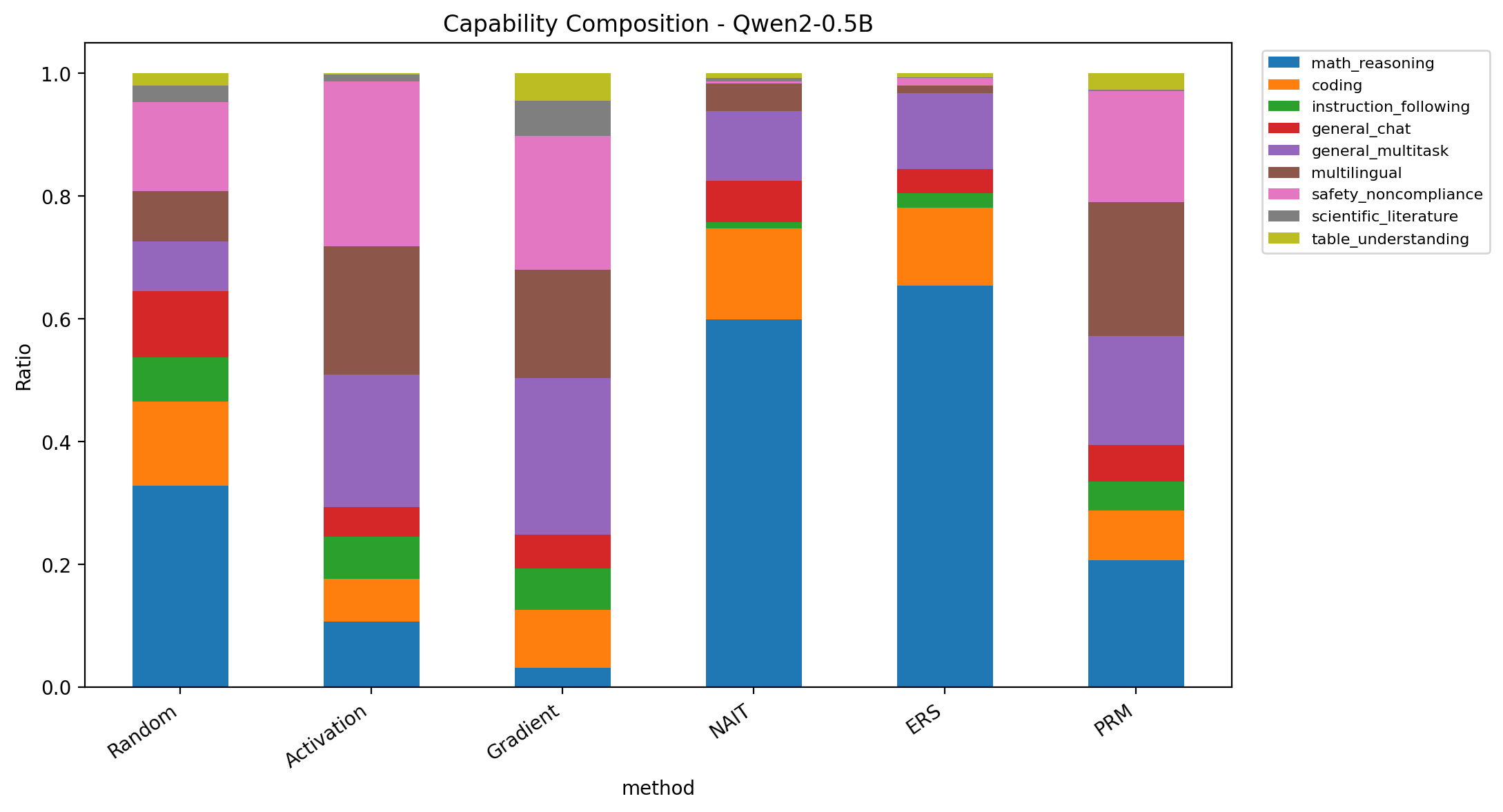}

\caption{
Capability composition of training subsets
selected by different methods on Qwen2-0.5B.
Each stacked bar represents the relative
proportions of nine capability categories
within the selected training subset.
}
\label{fig:capability_qwen05}
\end{figure*}

\begin{figure*}[t]
\centering
\includegraphics[
    width=0.95\textwidth
]{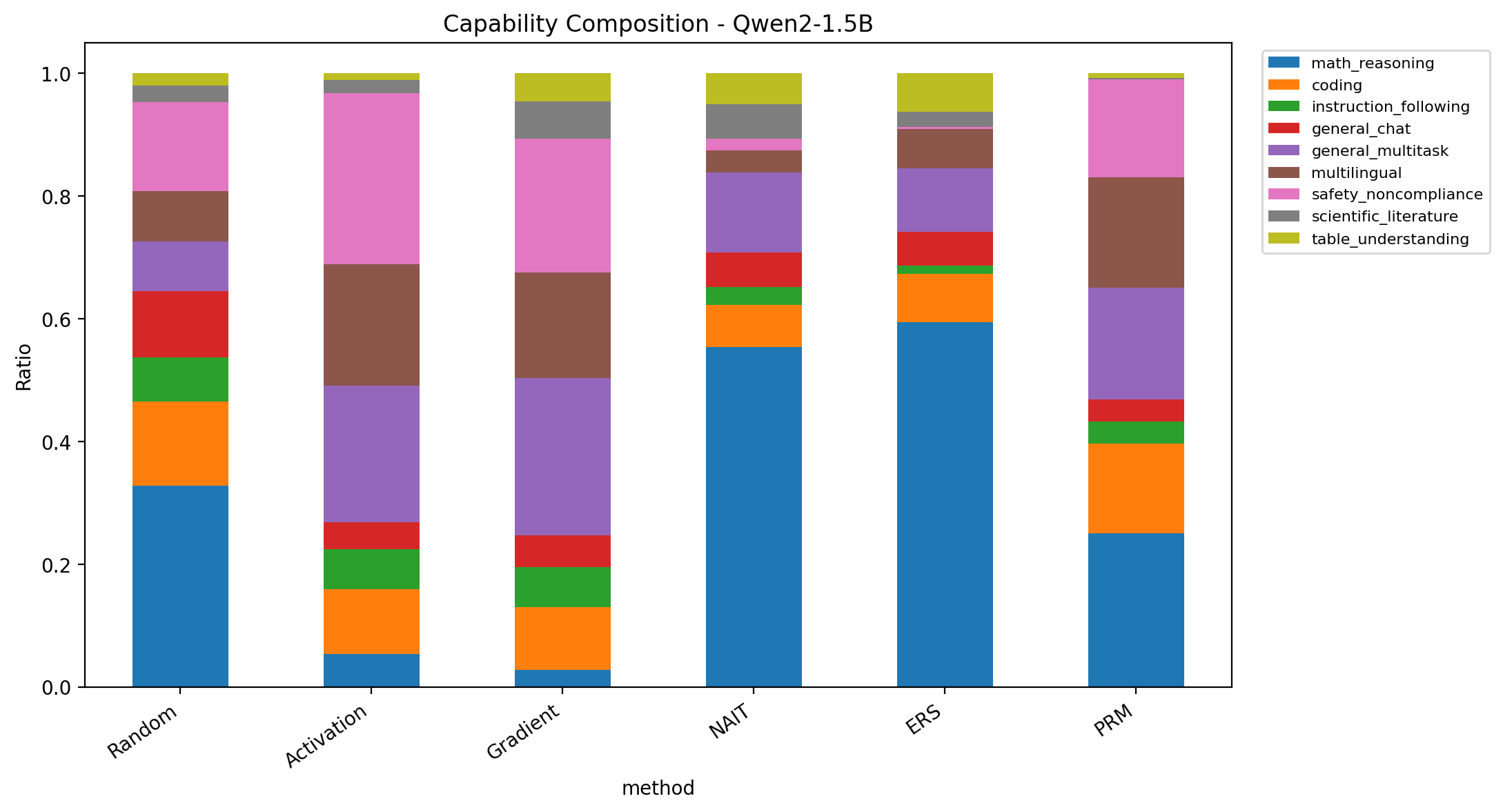}

\caption{
Capability composition of training subsets
selected by different methods on Qwen2-1.5B.
Each stacked bar represents the relative
proportions of nine capability categories
within the selected training subset.
}
\label{fig:capability_qwen15}
\end{figure*}

\begin{figure*}[t]
\centering
\includegraphics[
    width=0.95\textwidth
]{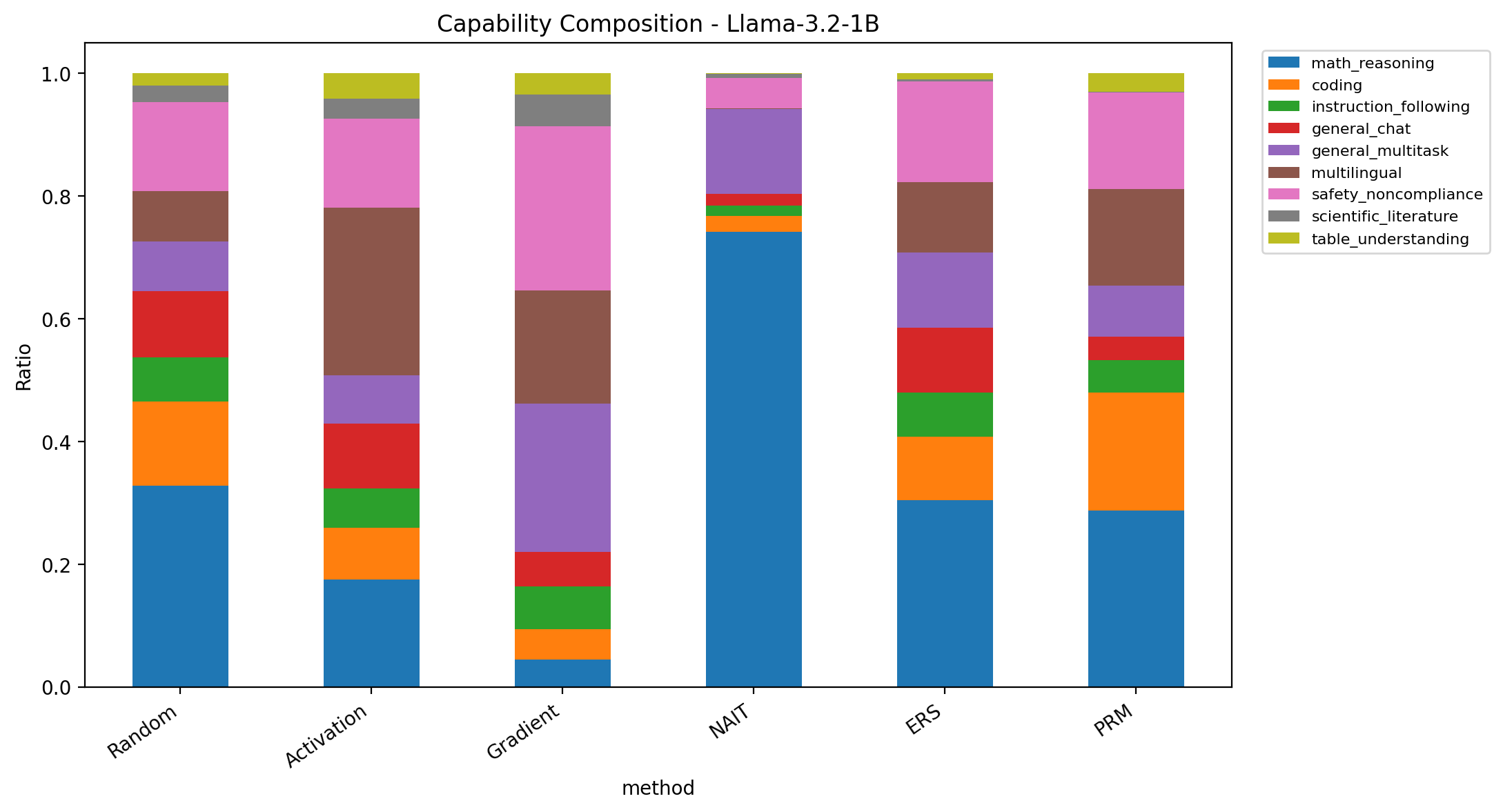}

\caption{
Capability composition of training subsets
selected by different methods on Llama-3.2-1B.
Each stacked bar represents the relative
proportions of nine capability categories
within the selected training subset.
}
\label{fig:capability_llama1b}
\end{figure*}

Figures~\ref{fig:capability_qwen05},
\ref{fig:capability_qwen15},
and~\ref{fig:capability_llama1b}
present the resulting capability distributions
for the three backbones.

These visualizations complement the
task-conditioned selection overlap analysis
by characterizing the task composition
of the selected training subsets.



\section{Answer Negative Log-Likelihood Distribution}
\label{sec:appendix_nll_distribution}

We further analyze the distribution of answer
negative log-likelihood (NLL) among training
examples selected by different methods.

Answer NLL measures the average negative
log-likelihood of answer tokens under the
corresponding scoring backbone.
Lower values indicate higher model-assigned
probability for the observed answers.

\begin{figure*}[t]
\centering
\includegraphics[
    width=0.85\textwidth
]{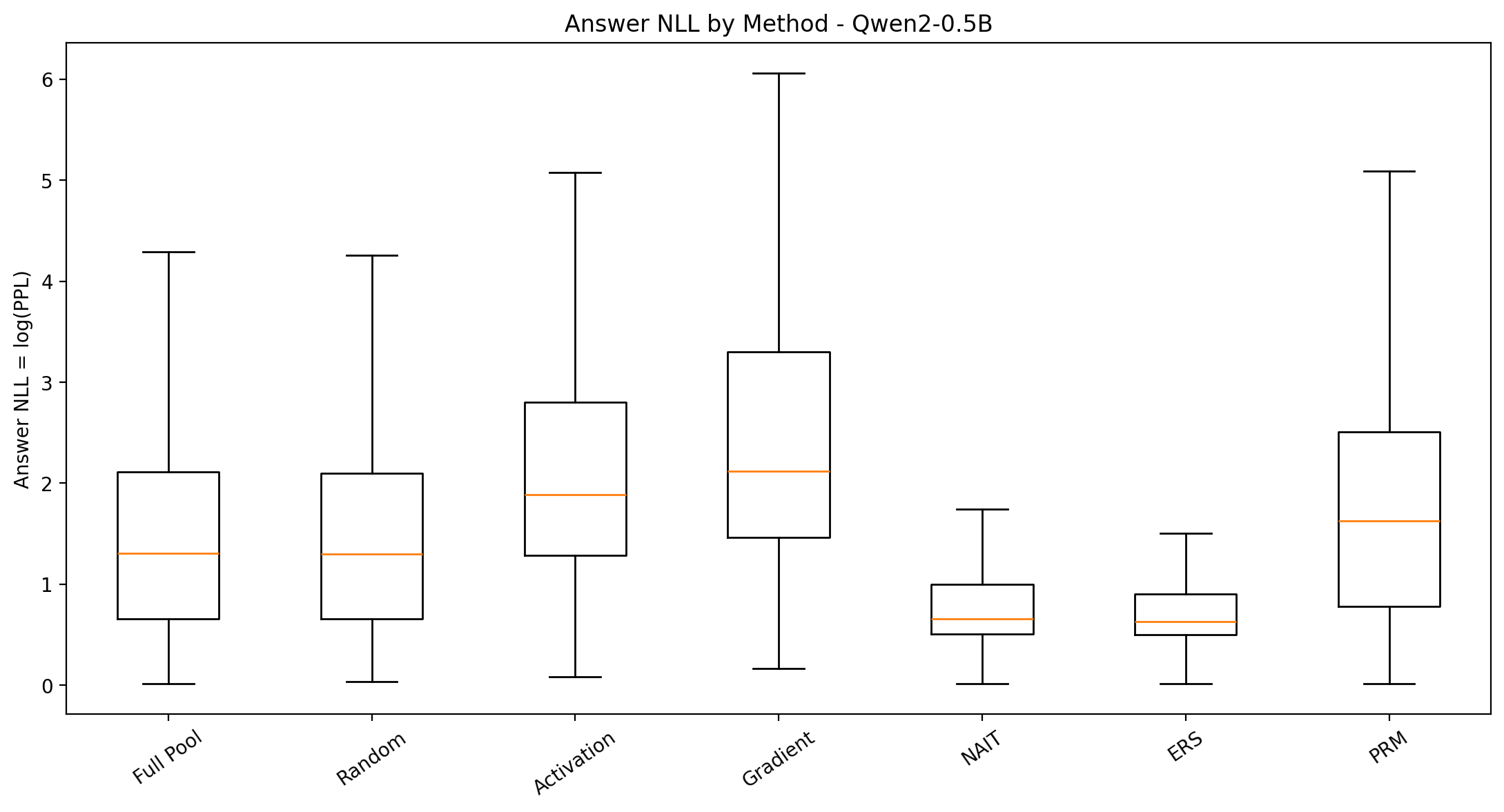}

\caption{
Answer NLL distributions of training examples
selected by different methods on Qwen2-0.5B.
The Full Pool distribution is included as
a reference.
}
\label{fig:nll_qwen05}
\end{figure*}

\begin{figure*}[t]
\centering
\includegraphics[
    width=0.85\textwidth
]{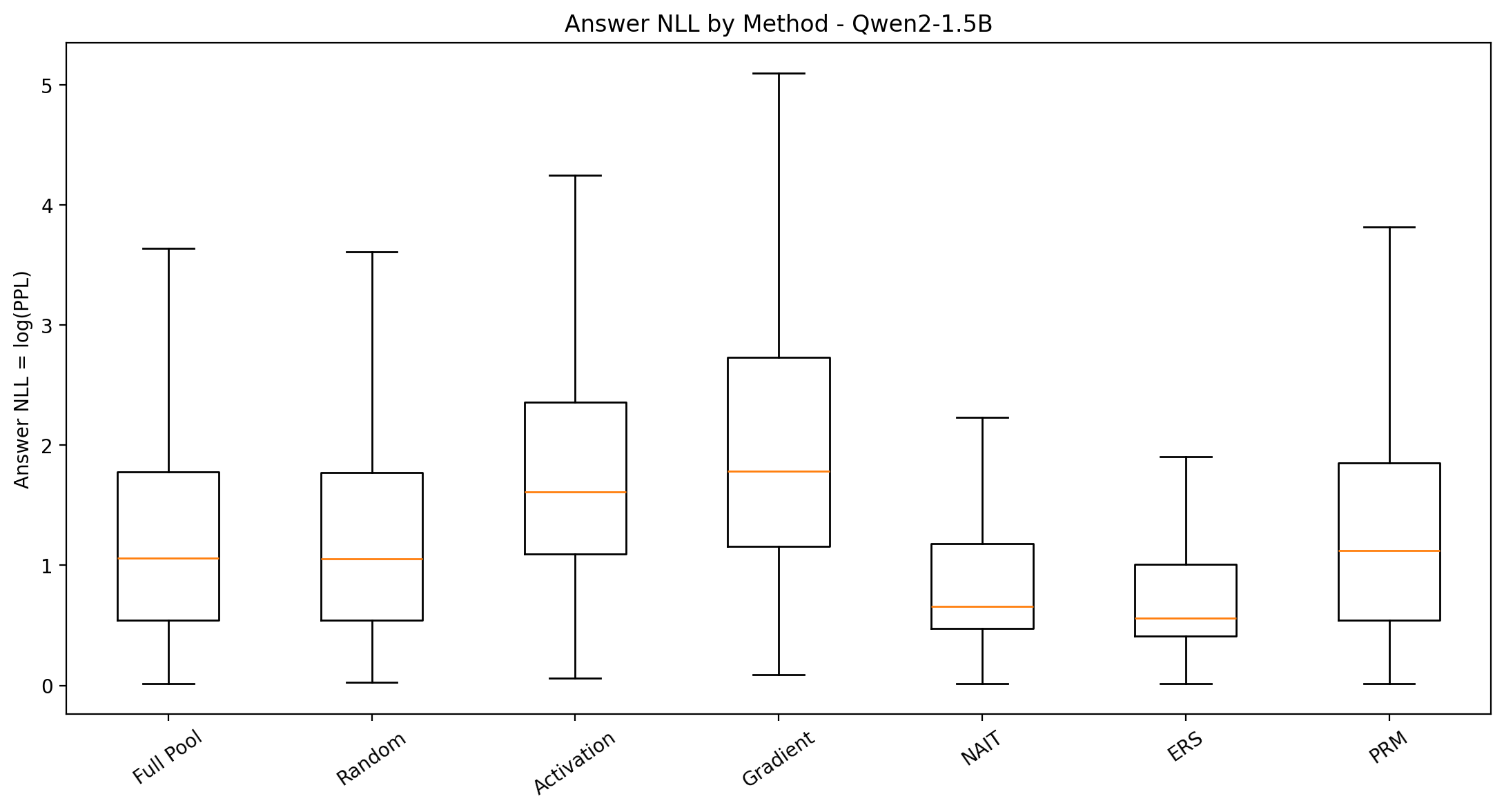}

\caption{
Answer NLL distributions of training examples
selected by different methods on Qwen2-1.5B.
The Full Pool distribution is included as
a reference.
}
\label{fig:nll_qwen15}
\end{figure*}

\begin{figure*}[t]
\centering
\includegraphics[
    width=0.85\textwidth
]{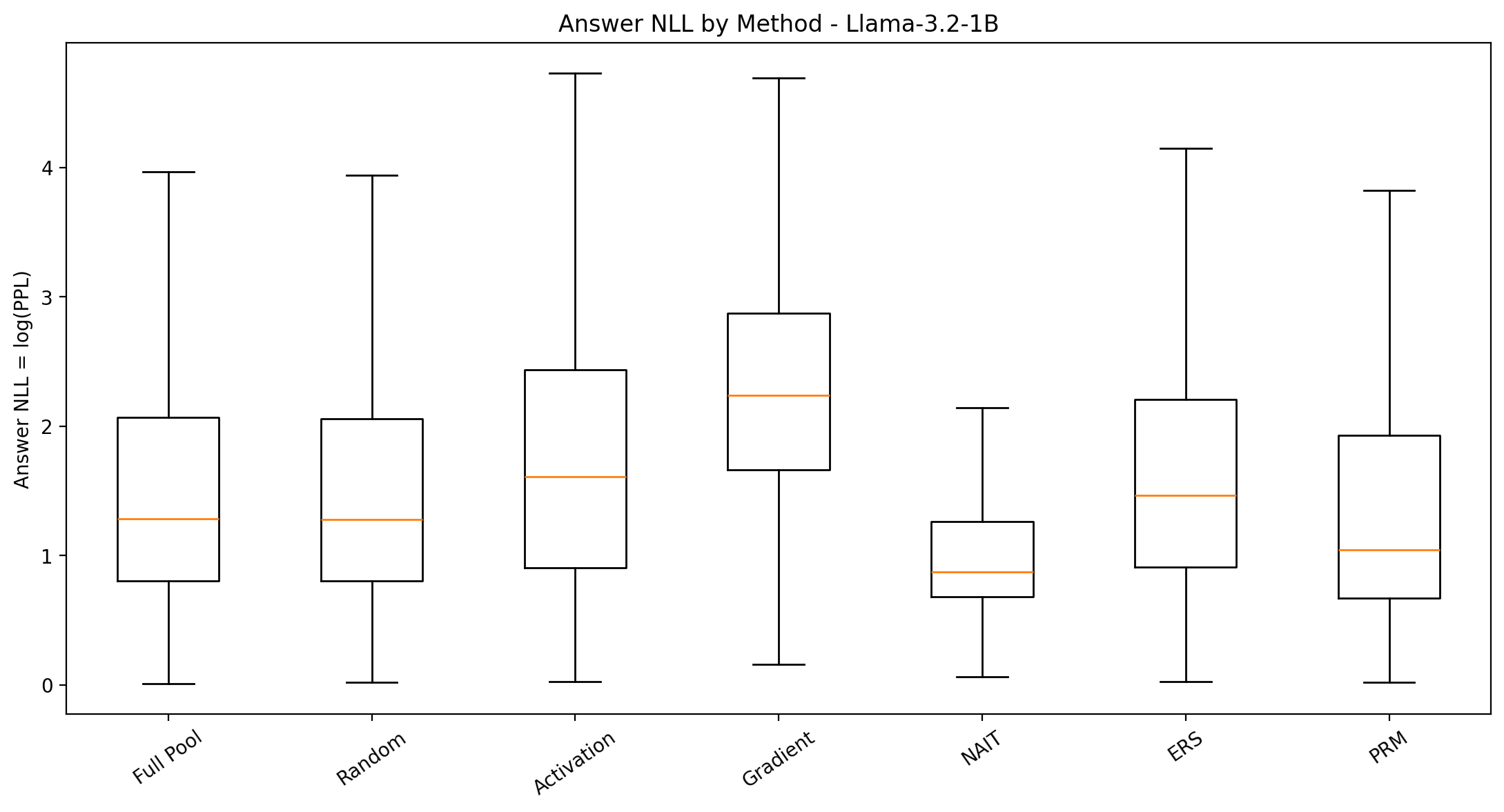}

\caption{
Answer NLL distributions of training examples
selected by different methods on Llama-3.2-1B.
The Full Pool distribution is included as
a reference.
}
\label{fig:nll_llama1b}
\end{figure*}

Figures~\ref{fig:nll_qwen05},
\ref{fig:nll_qwen15}, and
\ref{fig:nll_llama1b}
compare the answer NLL distributions of
selected examples against the full candidate
pool across three backbones.

These comparisons characterize the
likelihood-based preferences of different
data selection methods.


\section{Implementation Details of Data Selection Baselines}
\label{sec:appendix_baselines}

We provide additional implementation details
for the four baseline data selection methods:
Random, Activation, Gradient, and NAIT.
All methods operate on the same candidate
instruction pool with a fixed selection
budget of 30K examples.

\paragraph{Random.}

Random serves as a reference baseline
by uniformly sampling examples from
the candidate training pool.
It does not rely on model activations,
gradients, or external reference examples.

\paragraph{Activation.}

We adapt the activation probability measure
introduced in LAPE
\citep{tang-etal-2024-language}
to instance-level data selection.

For each instruction example $x$,
we measure the frequency of positive
FFN neuron activations across its tokens.

Let $a_{l,j,t}(x)$ denote the activation
of neuron $j$ in layer $l$ at token $t$.
Specifically, we use the output of the
SiLU-activated FFN gate projection.

The activation probability is defined as

\begin{equation}
p_{l,j}(x)
=
\frac{1}{T_x}
\sum_{t=1}^{T_x}
\mathbb{I}
\left[
a_{l,j,t}(x)>0
\right],
\end{equation}

where $T_x$ is the number of tokens
in example $x$.

We aggregate activation probabilities
across all FFN neurons and layers:

\begin{equation}
S_{\mathrm{act}}(x)
=
\frac{1}{LN}
\sum_{l=1}^{L}
\sum_{j=1}^{N}
p_{l,j}(x),
\end{equation}

where $L$ and $N$ denote the number
of Transformer layers and FFN neurons
per layer, respectively.

Examples are ranked by their
activation scores.

Unlike the original LAPE method,
which analyzes language-specific
activation patterns, our adaptation
uses the aggregated activation probability
as an instance-level data selection score.

\paragraph{Gradient.}

We adapt gradient-based neuron attribution
\citep{leng-xiong-2025-towards}
to instance-level data selection.

For each instruction example $x$,
we compute the language modeling loss
on its assistant response tokens.

Let $z_{l,j,t}(x)$ denote the FFN
intermediate activation of neuron $j$
in layer $l$ at token $t$.
We define its gradient-based attribution as

\begin{equation}
R_{l,j}^{\mathrm{grad}}(x)
=
\left|
\sum_{t=1}^{T_x}
z_{l,j,t}(x)
\frac{\partial \mathcal{L}(x)}
{\partial z_{l,j,t}(x)}
\right|,
\end{equation}

where $\mathcal{L}(x)$ is the
response-token language modeling loss.

Importantly, the absolute value
is applied after summing contributions
across tokens.

The instance-level selection score
is obtained by averaging attribution
magnitudes across all FFN neurons
and layers:

\begin{equation}
S_{\mathrm{grad}}(x)
=
\frac{1}{LN}
\sum_{l=1}^{L}
\sum_{j=1}^{N}
R_{l,j}^{\mathrm{grad}}(x).
\end{equation}

Examples are ranked according
to their gradient attribution scores.

This aggregation adapts neuron-level
gradient attribution to instance-level
data selection.

\paragraph{NAIT.}

We implement NAIT
\citep{chen2026neuron}
using reference-derived representation
directions for candidate scoring.

Unlike Activation and Gradient,
NAIT requires an external reference set
to construct its selection signal.

\textbf{Reference Set Construction.}
We construct a reference set containing
897 examples from three datasets:
ARC-Challenge, CommonsenseQA, and GSM8K,
with 299 examples from each dataset.

For ARC-Challenge, we use all 299 examples
from the validation split.
For CommonsenseQA and GSM8K,
we randomly sample 299 examples from
their respective training splits,
using a fixed random seed of 42.

The reference examples are drawn
from splits separate from those used
for downstream evaluation.
Specifically, evaluation uses the
ARC-Challenge test split,
CommonsenseQA validation split,
and test-side GSM8K examples
through MetaBench-GSM8K.

\textbf{Representation Direction Construction.}
Following NAIT, each reference example
is converted into a pair consisting
of its complete question-answer text
and a truncated version containing
only the first tokenizer token.

We extract the last-token hidden
representations of both sequences
at each Transformer layer.

Layer-wise representation differences
are used to fit a one-component PCA model,
yielding a representation direction
for each layer.

The orientation of each direction
is determined using the relative
projections of complete and truncated
reference examples.

\textbf{Candidate Scoring.}
For each candidate example,
we extract its last-token hidden
representation and project it
onto the reference-derived direction
at each layer.

The signed, centered projections
are averaged across layers to produce
the final NAIT selection score.

Examples are subsequently ranked
according to these scores.

Unlike NAIT, our proposed ERS
and PRM methods rely solely on
neuron relevance computed from
candidate training examples and
do not require target-task
reference data.


\section{Additional Training and Scoring Details}
\label{sec:appendix_training_details}

We provide the detailed training and
data-scoring configurations for the three
pretrained backbones in
Table~\ref{tab:training_config}.

All models are fine-tuned using the same
training framework and optimization strategy.
The training sequence length is adjusted
according to the backbone, while the
data-scoring sequence length is fixed.

\begin{table*}[t]
\centering
\small
\setlength{\tabcolsep}{7pt}
\renewcommand{\arraystretch}{1.15}

\begin{tabular}{lccc}
\toprule
\textbf{Configuration}
& \textbf{Qwen2-0.5B}
& \textbf{Qwen2-1.5B}
& \textbf{Llama-3.2-1B} \\
\midrule

\multicolumn{4}{l}{
\textit{Training Configuration}
} \\
\midrule

Training framework
& \multicolumn{3}{c}{LLaMA-Factory} \\

Fine-tuning strategy
& \multicolumn{3}{c}{Full-parameter SFT} \\

Candidate pool size
& \multicolumn{3}{c}{$\sim$100K} \\

Selected training examples
& \multicolumn{3}{c}{30K} \\

Optimizer
& \multicolumn{3}{c}{AdamW} \\

Learning rate
& \multicolumn{3}{c}{$2 \times 10^{-5}$} \\

LR scheduler
& \multicolumn{3}{c}{Cosine} \\

Warmup ratio
& \multicolumn{3}{c}{0.10} \\

Training epochs
& \multicolumn{3}{c}{3} \\

Sequence packing
& \multicolumn{3}{c}{Enabled} \\

Training sequence length
& 16,384
& 8,192
& 8,192 \\

\midrule
\multicolumn{4}{l}{
\textit{Data Scoring and Relevance Computation}
} \\
\midrule

Maximum scoring length
& 4,096
& 4,096
& 4,096 \\

Transformer layers
& 24
& 28
& 16 \\

Relevance computation
& \multicolumn{3}{c}{All Transformer layers} \\

\bottomrule
\end{tabular}

\caption{
Training and data-scoring configurations
for the three pretrained backbones.
Unless otherwise specified, configurations
are shared across models and selection methods.
}
\label{tab:training_config}
\end{table*}


\begin{figure*}[t]
\centering
\includegraphics[width=0.98\textwidth]{
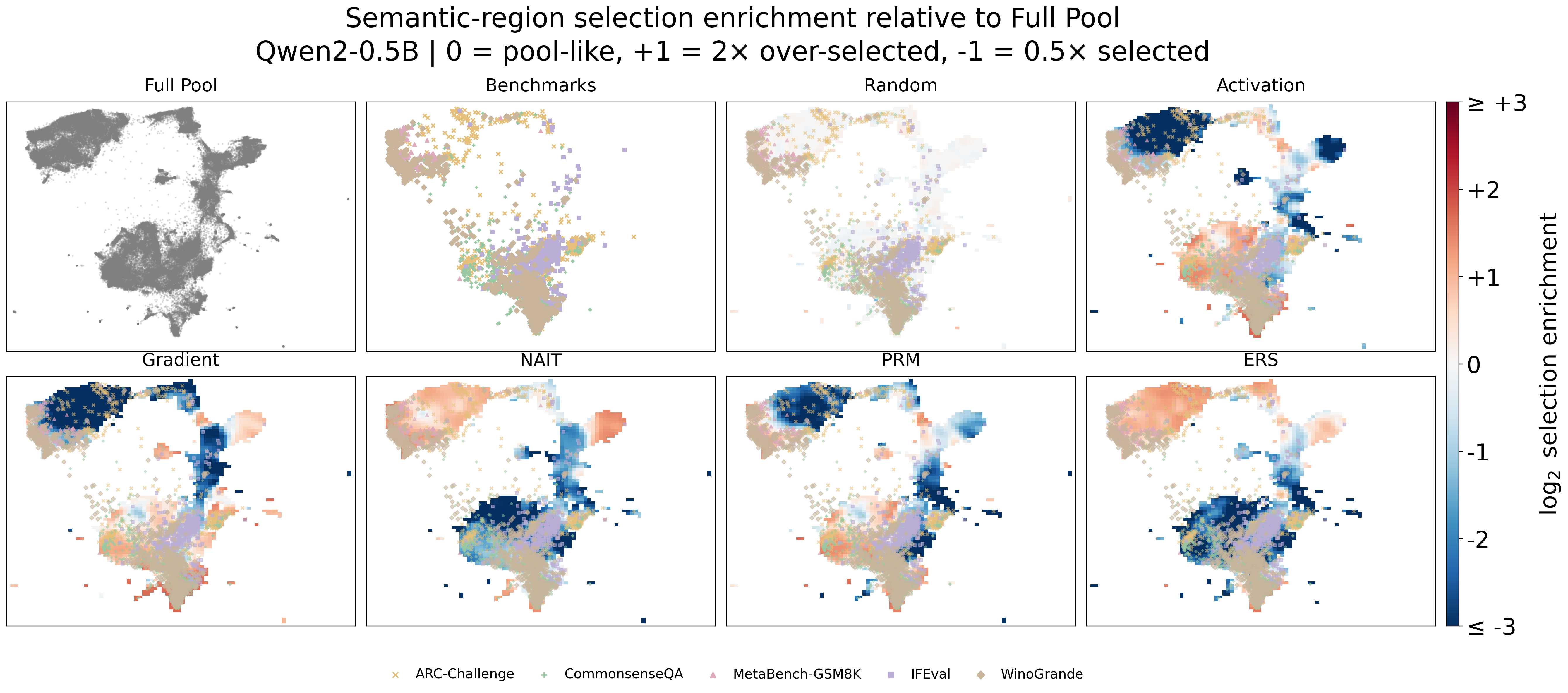
}
\caption{
Semantic-region selection enrichment on Qwen2-0.5B.
}
\label{fig:umap_qwen05}
\end{figure*}

\begin{figure*}[t]
\centering
\includegraphics[width=0.98\textwidth]{
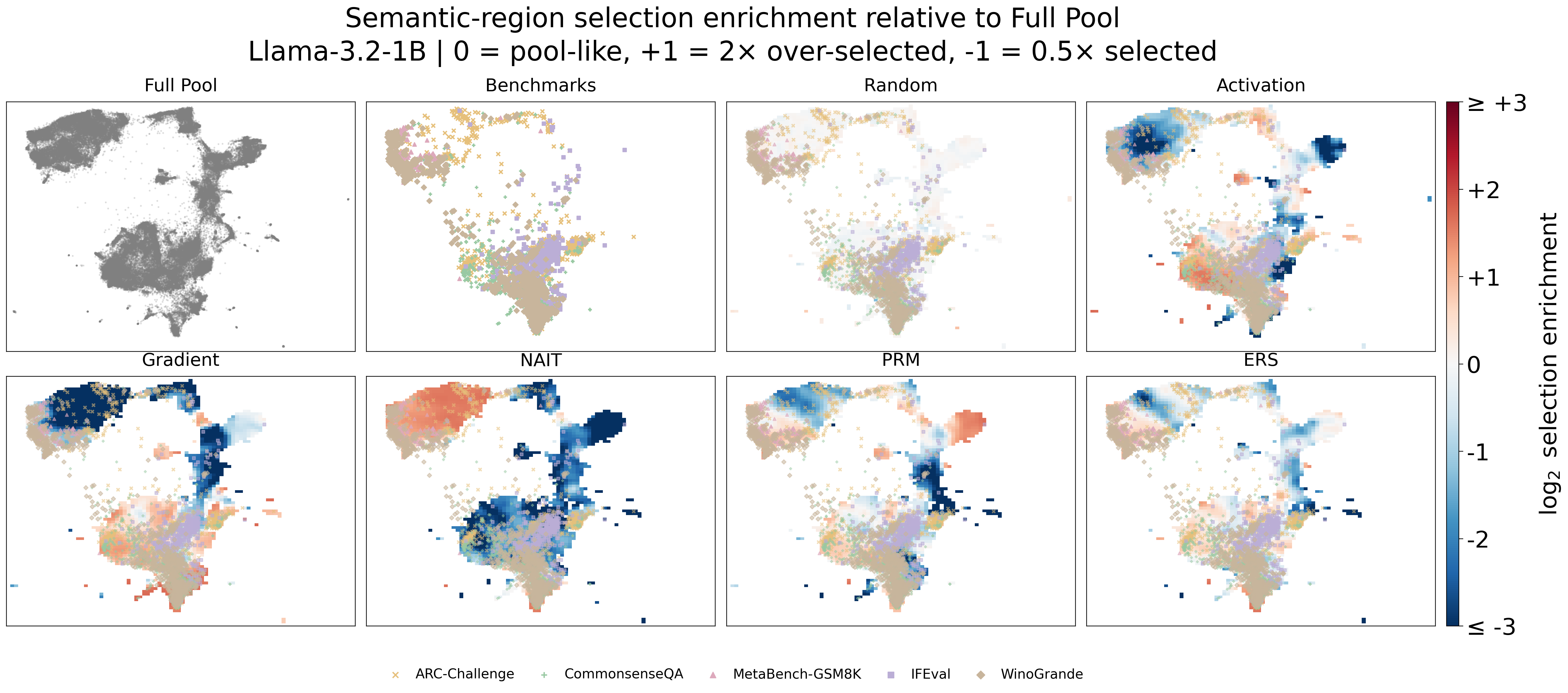
}
\caption{
Semantic-region selection enrichment on Llama-3.2-1B.
}
\label{fig:umap_llama1b}
\end{figure*}


\begin{figure*}[t]
\centering
\includegraphics[
    width=0.95\textwidth
]{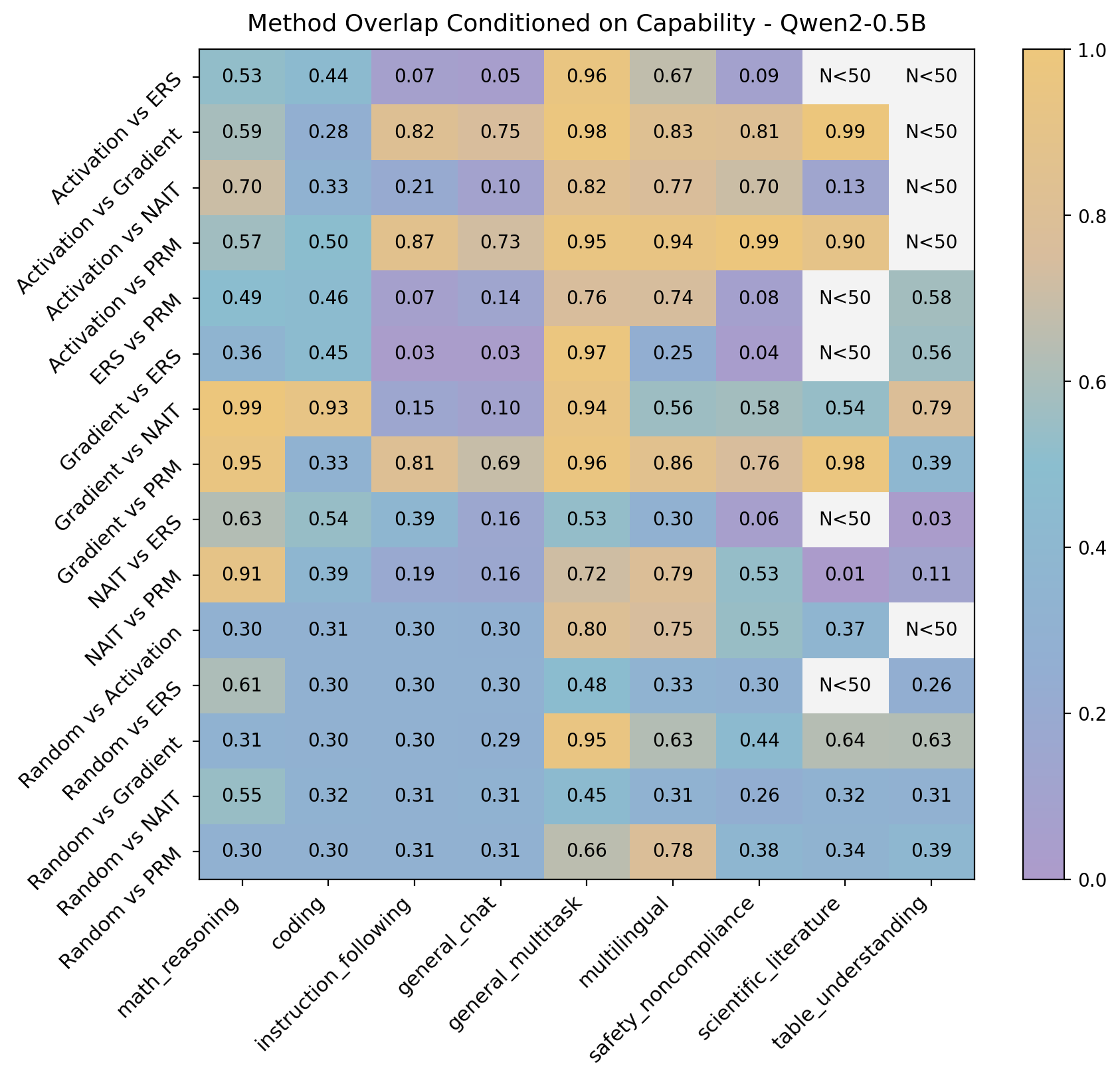}

\caption{
Task-conditioned selection overlap among
different methods on Qwen2-0.5B.
Overlap is normalized by the smaller selected
subset within each task category.
Cells with fewer than 50 selected examples
in either subset are masked.
}
\label{fig:task_overlap_qwen05}
\end{figure*}


\begin{figure*}[t]
\centering
\includegraphics[
    width=0.95\textwidth
]{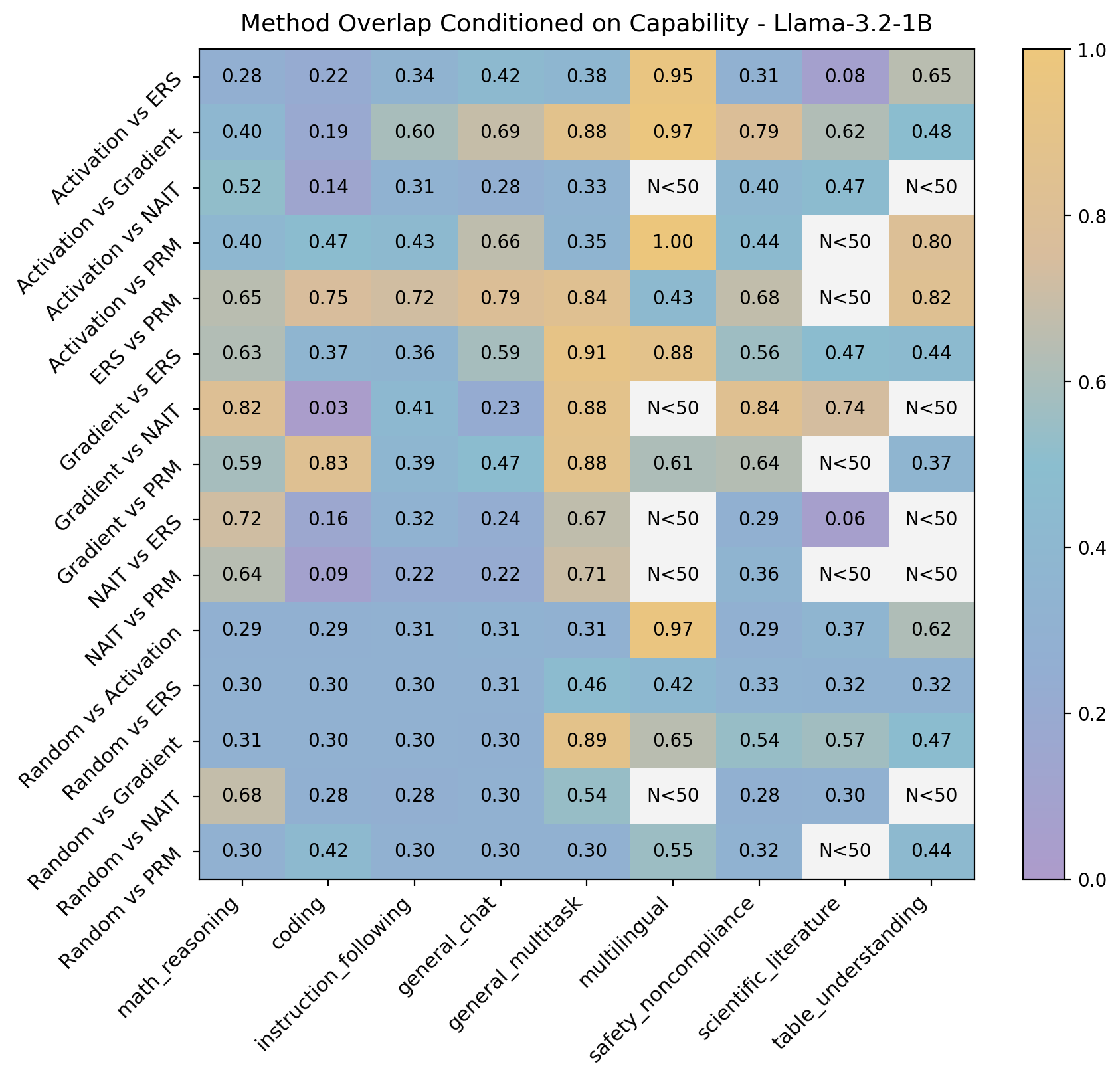}

\caption{
Task-conditioned selection overlap among
different methods on Llama-3.2-1B.
Overlap is normalized by the smaller selected
subset within each task category.
Cells with fewer than 50 selected examples
in either subset are masked.
}
\label{fig:task_overlap_llama1b}
\end{figure*}

\begin{table*}[t]
\centering
\small
\setlength{\tabcolsep}{6pt}
\begin{tabular}{lcccccc}
\toprule
\textbf{Task Type}
& \textbf{Random}
& \textbf{Activation}
& \textbf{Gradient}
& \textbf{NAIT}
& \textbf{ERS}
& \textbf{PRM} \\
\midrule

Math Reasoning
& 1.00 & 0.16 & 0.08 & 1.69
& \textbf{1.82} & 0.77 \\

Coding
& 1.00 & 0.77 & 0.75 & 0.50
& 0.57 & \textbf{1.06} \\

Instruction Following
& 1.01 & 0.90 & 0.92 & 0.41
& 0.20 & 0.50 \\

General Chat
& 1.01 & 0.41 & 0.48 & 0.52
& 0.52 & 0.33 \\

General Multitask
& 1.01 & 2.79 & \textbf{3.21} & 1.65
& 1.29 & 2.29 \\

Multilingual
& 0.99 & \textbf{2.36} & 2.06 & 0.42
& 0.76 & 2.13 \\

Safety Noncompliance
& 0.99 & \textbf{1.90} & 1.48 & 0.14
& 0.02 & 1.09 \\

Scientific Literature
& 1.00 & 0.82 & \textbf{2.31} & 2.09
& 0.93 & 0.09 \\

Table Understanding
& 0.95 & 0.52 & 2.20 & 2.45
& \textbf{3.05} & 0.37 \\

\bottomrule
\end{tabular}
\caption{
Task-type enrichment of selected data on
Qwen2-1.5B.
Values indicate the ratio between the proportion
of a task category in the selected subset and
its proportion in the full candidate pool.
Values above one indicate over-representation.
}
\label{tab:task_enrichment}
\end{table*}

\providecommand{\topone}[1]{%
    \textbf{#1}\textsuperscript{\scriptsize 1}}
\providecommand{\toptwo}[1]{%
    #1\textsuperscript{\scriptsize 2}}
\providecommand{\topthree}[1]{%
    #1\textsuperscript{\scriptsize 3}}


\begin{table*}[t]
\centering
\small
\setlength{\tabcolsep}{5pt}
\renewcommand{\arraystretch}{1.08}

\begin{tabular}{llccccc@{\hspace{10pt}}c}
\toprule

\textbf{Backbone}
& \textbf{Method}
& \textbf{ARC-C}
& \textbf{CSQA}
& \textbf{Meta-GSM8K}
& \textbf{IFEval}
& \textbf{WinoGrande}
& \textbf{Avg. Rank} $\downarrow$ \\

\midrule


\multirow{7}{*}{Llama-3.2-1B}
& \textit{Base}
& 39.68 & 52.01 & 6.75 & 10.54 & 60.30
& -- \\

\cmidrule(lr){2-8}

& Random
& 40.22
& 49.39
& \topone{14.77}
& \topone{16.14}
& \toptwo{63.56}
& \topone{2.80} \\

& Activation
& 40.13
& \topthree{50.18}
& 10.97
& \topthree{15.47}
& 62.83
& \topthree{4.40} \\

& Gradient
& \topthree{40.59}
& 48.95
& 10.97
& \toptwo{15.77}
& 62.75
& 4.60 \\

& NAIT
& 40.36
& 49.74
& \topthree{13.64}
& 10.60
& \topone{63.59}
& \toptwo{3.60} \\

& \textbf{PRM}
& \topone{41.61}
& \toptwo{50.34}
& \toptwo{13.78}
& 13.62
& 62.88
& \topone{2.80} \\

& \textbf{ERS}
& \toptwo{40.73}
& \topone{50.70}
& 11.53
& 14.60
& \topthree{63.30}
& \topone{2.80} \\

\midrule


\multirow{7}{*}{Qwen2-0.5B}
& \textit{Base}
& 31.06 & 50.78 & 36.29 & 13.12 & 58.09
& -- \\

\cmidrule(lr){2-8}

& Random
& \topthree{31.40}
& \topone{50.64}
& 36.15
& \topone{15.22}
& \toptwo{56.41}
& \toptwo{2.20} \\

& Activation
& \topthree{31.40}
& 49.00
& \toptwo{37.97}
& 13.80
& 56.20
& 4.00 \\

& Gradient
& \toptwo{31.66}
& 49.66
& 31.93
& \toptwo{14.79}
& 55.56
& 4.20 \\

& NAIT
& 31.26
& \topthree{50.12}
& 35.02
& 12.75
& \topthree{56.22}
& 4.20 \\

& \textbf{PRM}
& 31.20
& 49.90
& \topone{39.38}
& 13.99
& 55.99
& \topthree{3.80} \\

& \textbf{ERS}
& \topone{32.08}
& \toptwo{50.45}
& \topthree{36.85}
& \topthree{14.66}
& \topone{56.41}
& \topone{2.00} \\

\midrule


\multirow{7}{*}{Qwen2-1.5B}
& \textit{Base}
& 44.45 & 67.40 & 59.49 & 16.08 & 66.38
& -- \\

\cmidrule(lr){2-8}

& Random
& 42.83
& \topthree{69.18}
& \topone{60.62}
& 18.30
& 65.32
& 3.60 \\

& Activation
& \topthree{43.00}
& 68.96
& 57.38
& \toptwo{19.65}
& \topone{66.22}
& \toptwo{2.80} \\

& Gradient
& \topone{43.77}
& \topone{69.37}
& 54.43
& 16.76
& \toptwo{65.98}
& \topthree{3.20} \\

& NAIT
& 42.58
& \topthree{69.18}
& 55.70
& 19.53
& \toptwo{65.98}
& 4.00 \\

& \textbf{PRM}
& \toptwo{43.17}
& \toptwo{69.31}
& \topthree{58.09}
& \topthree{19.59}
& \topthree{65.80}
& \topone{2.60} \\

& \textbf{ERS}
& 42.80
& 68.20
& \toptwo{59.92}
& \topone{20.02}
& 65.59
& 3.40 \\

\bottomrule

\end{tabular}

\caption{
Mean downstream performance (\%) across three
backbones and five benchmarks over three training
epochs.
Superscripts $1$, $2$, and $3$ indicate the
top three methods, with the best highlighted
in bold.
Avg. Rank is computed across the five benchmarks
(lower is better).
\textit{Base} is excluded from ranking.
}

\label{tab:mean_main_results}

\end{table*}

\end{document}